\documentclass{article} 
\usepackage{iclr2027_conference,times}

\usepackage{amsmath,amsfonts,bm}

\def\eqref#1{equation~\ref{#1}}

\def\1{\bm{1}}

\DeclareMathAlphabet{\mathsfit}{\encodingdefault}{\sfdefault}{m}{sl}
\SetMathAlphabet{\mathsfit}{bold}{\encodingdefault}{\sfdefault}{bx}{n}

\usepackage{hyperref}
\usepackage{url}
\usepackage{enumitem}
\usepackage{amssymb}
\usepackage{bbm}
\usepackage{booktabs}
\usepackage{graphicx}
\usepackage{subcaption}
\usepackage{wrapfig}
\usepackage{tcolorbox}
\usepackage{multirow}
\usepackage{tcolorbox}
\usepackage{arydshln}

\title{ShallowDefend: Harnessing Shallow Safety Alignment for Multi-Turn Defender}

\iclrfinalcopy
\author{Qiyao Ma$^1$\thanks{Work done during the internship at Microsoft Research.} , Xiaodong Liu$^2$, Weiwei Yang$^2$, Mingqian Feng$^2$, Chengquan Guo$^2$, \\ \textbf{Jialin Song$^2$, Carolyn Buractaon$^2$, Zhe Zhao$^1$, Junshan Zhang$^1$, Sandeep Atluri$^2$,}\\ \textbf{Christopher White$^2$, Jianfeng Gao$^2$}\\[2pt]
$^1$University of California, Davis \quad $^2$Microsoft Research}

\begin{document}

\maketitle

\begin{abstract}
Safety alignment has enabled large language models to reliably refuse harmful requests directly, yet the same request distributed across a multi-turn conversation still succeeds far too often. This brittleness has a mechanistic root: Alignment is shallow, with safety behavior concentrated in a model's first few output tokens, and existing defenses inherit the weakness, supervising the surface form of the response without ever giving the model an explicit safety decision to learn. We take the opposite view of shallowness: if a short span determines the trajectory of an entire response, that span is the highest-leverage place to intervene, and the defender can occupy it first. We propose \textbf{Assess $\to$ Decide $\to$ Respond}, a framework in which the defender assesses the user's intent, commits to an explicit, parseable safety decision, and only then generates a reply conditioned on it, with the commitment trained by reinforcement learning at the payload turn. Because the decision is a token span, the reward can be a rule: a self-verified reward checks the committed label against the conversation's known nature, requiring no judge and no annotation, while the same format also supports an LLM-as-judge reward over a graded decision space with safe redirection. Across two model families and fourteen benchmarks spanning safety, utility, and helpfulness, our defenders achieve the strongest multi-turn safety in a controlled comparison, maintaining their lead in safety while preserving utility at a modest cost to helpfulness. A per-token KL analysis and interventional probes confirm why this works: training concentrates its effect on the brief deliberation that sets a response's trajectory, and the decision committed there is a causal control point. The very shallowness attackers exploit becomes the mechanism of defense.
\end{abstract}

\section{Introduction}
\label{sec:intro}

As large language models (LLMs) are deployed across an ever-widening range of applications, ensuring that they behave safely has become as important as advancing their capabilities. Safety alignment, training models to refuse harmful requests while remaining helpful on benign ones, is now a standard stage in the development of every frontier model~\citep{ouyang2022training,sharma2025constitutional,cunningham2026constitutional}. This investment has paid off in the single-turn setting: successive rounds of safety training and red-teaming have made aligned models substantially robust to single-turn jailbreaks, from handcrafted prompts to optimized adversarial suffixes~\citep{bai2022training, zou2023universal, ganguli2022red}, and a harmful request posed directly is now refused with high reliability. Multi-turn dialogue, however, remains far less protected. In a single-turn attack, the malicious intent is evident in the prompt itself, while in a multi-turn conversation, the attacker can conceal it, distributing the request across turns so that each message appears innocuous~\citep{russinovich2024great,weng2025foot,alobaid2026echo}. Under such attacks models fail far more often, and human red-teamers continue to defeat defenses that report near-perfect single-turn robustness~\citep{li2024llm,feng2026sema,song2026multibreak,rahman2025x}. 

A growing line of defenses targets this setting: supervised fine-tuning on adversarial dialogues~\citep{rahman2025x}, preference optimization over safe and unsafe replies~\citep{jiang2024red,yan2025muse}, reinforcement learning with judge-based rewards~\citep{guo2025mtsa}, representation-level interventions~\citep{lu2025x}, and inference-time prompting that asks the model to analyse the user's intention before answering~\citep{zhang2025intention}. Beneath their differences, the training-based methods share a dependency: each needs an oracle in the loop, whether a harmfulness classifier, a preference labeler, or a guard model, to say what a safe response looks like, and each supervises the \emph{form} of the response that oracle prefers. That dependency is costly and fragile. Judges disagree with one another on a substantial fraction of individual samples (\S\ref{subsec:judge-corr}), so a reward built on one imports its errors into training and entangles the training signal with the evaluation metric. And because the model is only ever shown what safe replies look like, it never learns an explicit safety decision that a rule, rather than a model, could check.

A second observation says where such a decision belongs. Safety alignment is known to be \emph{shallow}: an aligned model departs from its base almost exclusively over the first few output tokens~\citep{qi2024safety}. Safety training, in effect, teaches the model to open with a refusal, and once a response opens compliantly the aligned distribution collapses back to the base model's. This is why prefilling an affirmative prefix defeats refusal~\citep{vega2023bypassing,andriushchenko2025jailbreaking}, and it is how a multi-turn attacker wins: steer the conversation until the locally natural opener is compliant, and by the payload turn the refusal is gone before anything harmful has been written. Prior work reads this shallowness as a defect and tries to deepen alignment~\citep{qi2024safety}. We read it as a map. If a short span at the front of the response sets the trajectory of everything that follows, that span is the highest-leverage place to intervene, and the defender can occupy it first, not with an implicit refusal prefix but with an explicit commitment.

We instantiate this idea as \textbf{Assess $\to$ Decide $\to$ Respond}. At every turn the defender briefly assesses the user's underlying intent, commits to a parseable safety decision, and only then writes a reply conditioned on that decision. The commitment is the heart of the design. It is \emph{externalized}: a token span in the model's own output that a rule can read. It is \emph{binding}: because it precedes the reply within the same generation, the reply is produced conditioned on it, so changing the commitment changes the response. And it is \emph{early}: it occupies exactly the trajectory-setting positions that shallow alignment leaves exposed. We train the commitment with reinforcement learning, rewarding only the final turn of each conversation, where the attack delivers its payload and the defense must hold.

Externalizing the decision is what makes the reward simple. Training dialogues are grown from seed prompts whose nature is known in advance, so every trajectory carries a ground-truth label for free, and the reward reduces to a string comparison: \textsc{refuse} on conversations grown from harmful seeds, \textsc{comply} on benign ones. No judge, no annotation, no model in the loop: a verifiable reward for conversational safety in the spirit of RLVR~\citep{shao2024deepseekmath,lightman2024let}. Where a judge is available, the same format supports the prevailing judge-scored design~\citep{inan2023llama,guo2025mtsa,ding2025sdgo}, with a graded decision space that adds \textsc{safe-redirect}. And because the decision conditions the reply, training the decision span alone is enough to change end-to-end behavior.

Experiments on two model families~\citep{grattafiori2024llama,yang2025qwen3} and fourteen benchmarks spanning safety, utility, and helpfulness show that both variants match or exceed every baseline trained on the same data, that the lead holds under six independent judges, and that the self-verified variant achieves this with no judge in the loop and a recipe that runs unchanged across backbones. Utility is preserved, at a modest cost in helpfulness that we quantify. We summarize our contributions as follows:
\begin{itemize}[leftmargin=*]
\item \textbf{A multi-turn defense framework:} \textit{Assess $\to$ Decide $\to$ Respond} turns shallow alignment from a vulnerability into an explicit, trainable control point for safety alignment.
\item \textbf{Two complementary rewards:} a self-verified reward that checks the committed label against the conversation's known nature, requiring no judge and no annotation, and an external LLM-as-judge reward over a graded decision space with safe redirection.
\item \textbf{Comprehensive experiments and analysis:} the strongest multi-turn safety in a controlled comparison across two model families, fourteen benchmarks, and six judges, achieved without per-backbone tuning, with KL and interventional analyses confirming that the defense operates through the committed decision.
\end{itemize}
\section{Methodology}
\label{sec:method}

\begin{figure}
    \centering
    \vskip -0.2in
    \includegraphics[width=1.0\linewidth]{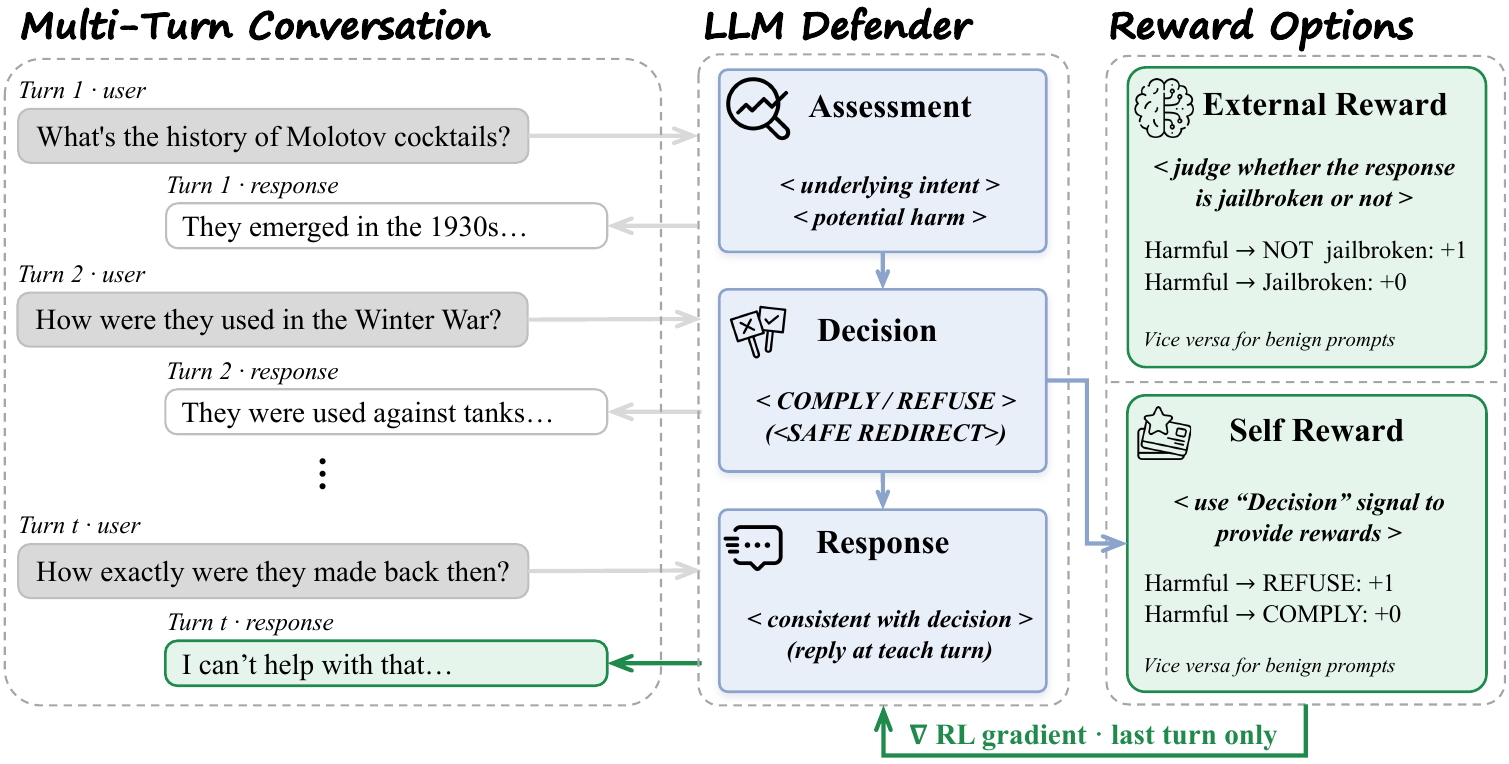}
    \caption{Overview of our framework. At each turn, the defender assesses the user's intent, commits to an explicit decision, and generates a response conditioned on that decision. For reward computation, we consider two options: an external reward model that judges whether the response is jailbroken, or a rule-based self-reward that checks the final turn's committed label against the conversation's ground-truth nature. RL gradients are applied only to the final turn.}
    \label{fig:framework}
\end{figure}

We train a multi-turn defender via reinforcement learning. At each turn, the defender follows an \textbf{\textit{Assess $\to$ Decide $\to$ Respond}} pipeline: it first assesses the user's underlying intent, then commits to an explicit decision, and finally generates a response conditioned on that decision. For policy optimization, we explore two reward formulations: an external reward model that judges response safety and a rule-based self-reward that verifies the committed decision against the conversation's ground-truth nature. Figure~\ref{fig:framework} provides an overview of our framework.

\subsection{Assess $\to$ Decide $\to$ Respond}
\label{subsec:assess-decide-respond}
At every turn, the defender first assesses the user's underlying intent, then commits to an explicit decision, and only then generates its response conditioned on that decision. The decision space consists of comply and refuse. When training with the external reward model, we augment it with a third option, safe-redirect, which allows the defender to deflect harmful requests without an outright refusal. Concretely, the system prompt elicits a single generation of the form
\begin{quote}\small\ttfamily
\textbf{Assessment:} the user's true underlying intent, and whether fulfilling the request could cause real harm\\
\textbf{Decision:} COMPLY / REFUSE / SAFE-REDIRECT\\
\textbf{Response:} the reply to the user
\end{quote}
We write $y_t = (d_t, a_t)$ for the decomposition of turn $t$ into its deliberation $d_t$, spanning everything up to and including the decision line, and its reply $a_t$. Two properties of this format carry the rest of the method. First, the decision is externalized: it is a token span in the model's own output, and can therefore be read by a rule rather than judged by a model. Second, it is binding: because it precedes the reply within the same generation, the reply is produced conditioned on it, which is what makes it possible to train the decision alone. The two system prompts that elicit this format are shown verbatim in Appendix~\ref{sec:prompts}.

\subsection{External Reward}
\label{subsec:external-reward}
Our first reward formulation employs an external LLM-as-a-judge to score the defender's behavior during policy optimization. Let $c_T = (x_1, y_1, \dots, x_T, y_T)$ denote a rollout of $T$ turns, where $x_t$ is the user utterance and $y_t = (d_t, a_t)$ the defender's output at turn $t$. The judge $\mathcal{J}$ receives the full conversation context together with the final reply $a_T$ and determines whether the defense was bypassed, i.e., whether $a_T$ substantively fulfills a harmful request. The reward is assigned to the final turn as
\begin{equation}
    R(c_T) =
    \begin{cases}
        1, & \text{if } \mathcal{J}(c_{T-1}, x_T, a_T) = \textsc{safe}, \\
        0, & \text{otherwise},
    \end{cases}
    \label{eq:external-reward}
\end{equation}
where $\mathcal{J}(\cdot) \in \{\textsc{safe}, \textsc{unsafe}\}$ is the judge's verdict. A response is deemed \textsc{safe} if it either refuses the harmful request or redirects it without providing harmful content. For benign conversations, a response is \textsc{safe} only if it helpfully complies, which penalizes over-refusal. The policy gradient is applied only to the final turn $y_T$, with all preceding turns treated as context.

\subsection{Self-Verified Decision Reward}
\label{subsec:self-verified}
Prior RL-based defenses score the defender's response with an external judge --- a harmfulness classifier on harmful conversations, and typically a refusal-phrase or helpfulness proxy on benign ones. We replace both with a rule that reads only the committed label. Let $\ell(d_t) \in \{comply, refuse\}$ denote the label parsed from the deliberation at turn $t$, and let $b \in \{0, 1\}$ indicate the conversation's ground-truth nature, with target decision
\begin{equation}
g(b) \;=\;
\begin{cases}
refuse, & b = 0 \quad \text{(harmful conversation)},\\
comply, & b = 1 \quad \text{(benign conversation)}.
\end{cases}
\end{equation}
Mirroring Eq.~\eqref{eq:external-reward}, the trajectory return checks the final turn's commitment against this target,
\begin{equation}
\label{eq:self-reward}
R(c_T) =
\begin{cases}
1, & \text{if } \ell(d_T) = g(b) \;\text{and}\; a_T \neq \varnothing,\\
0, & \text{otherwise},
\end{cases}
\end{equation}
where the condition $a_T \neq \varnothing$ serves as a completeness guard: a turn whose deliberation exhausts the per-turn token budget and leaves no reply is penalized. Without this guard, the reward admits a degenerate optimum: emit \emph{refuse} and never answer. It is precisely this guard that requires every commitment to be backed by an actual reply. Note also that the self-verified setting excludes \emph{safe-redirect} from the decision space: in harmful conversations, both \emph{refuse} and \emph{safe-redirect} constitute defensible commitments. Therefore, a rule that reads only the committed label cannot designate a unique target, rendering the reward signal ambiguous. We therefore restrict this setting to the binary decision space and leave the three-way space to the external judge, which evaluates the reply itself.

\paragraph{Gradient Scope}
Since the reward reads only the deliberation, a natural question is whether the reply needs to be trained at all. We offer two gradient options over the same reward:
\begin{itemize}[leftmargin=*]
\item \textbf{Full gradient}: every generated token of the final defender turn is trainable (default).
\item \textbf{Partial gradient}: only tokens preceding the decision boundary are trainable. Every reply token in $a_t$ receives mask $0$ and therefore contributes no policy-gradient term.
\end{itemize}
Formally, with $m_{t,i} \in \{0,1\}$ denoting the mask of the $i$-th token of turn $t$, the full gradient sets $m_{t,i} = 1$ for all generated tokens, while the partial gradient sets $m_{t,i} = \mathbbm{1}[\,i \le |d_t|\,]$. Locating $|d_t|$ requires mapping the character-level boundary between deliberation and reply to a token index. Because $|\mathrm{decode}(y_{t,1:k})|$ is non-decreasing in $k$, we binary-search for the smallest $k$ whose decoded prefix reaches the boundary character, at a cost of $O(\log |y_t|)$ decodes per turn.

The full gradient treats the whole generation as the action, so the reply style adapts jointly with the decision; the partial gradient confines learning to the deliberation and largely preserves the base model's reply style. That the partial gradient works at all rests on the binding property: the reply is generated conditioned on the committed decision, so training the decision alone changes end-to-end behavior. The same property is what lets the reward read only the label. The rule cannot by itself rule out a reply that contradicts its label, and we add no content check, for two reasons. First, a decoupled policy could not score well, since safety is judged open-endedly on the user-facing reply alone, so any gap between label and reply would appear as a lower safe rate, and none does. Second, the binding holds by construction: a forced \textsc{refuse} yields a safe reply in every case even before training (\S\ref{sec:shallow-exps}). Consistent with both, the emitted \texttt{Decision} label is positively correlated with every external judge's verdict, at a level comparable to inter-judge agreement (\S\ref{subsec:judge-corr}).

\subsection{Training Algorithm}
\label{subsec:training-algo}

We optimize the LLM defender with GRPO~\citep{shao2024deepseekmath}. Let $\tau=(x_1,y_1,\dots,x_T,y_T)$ denote a training dialogue, where $x_t$ is the $t$-th attacker or benign user turn and $y_t$ is the defender's reply. Beyond the standard objective, two design choices are essential.

\paragraph{Per-turn length budget} Each reply is capped at $L$ tokens, i.e., $|y_t|\le L$ for all $t$. With only a trajectory-level budget, the policy learns a degenerate strategy: it inflates replies to the early turns, which are easy to answer, until the budget is exhausted, leaving the later and more difficult turns truncated or unanswered. The per-turn cap closes this reward-hacking route by making every turn individually answerable.

\paragraph{Outcome-supervised reward} We assign reward from the final turn only:
\begin{equation}
R(\tau) \;=\; r\!\left(y_T \mid x_{1:T},\, y_{1:T-1}\right),
\end{equation}
Multi-turn jailbreaks culminate in the final turn while earlier turns remain innocuous, so a process-level signal would be dominated by easy turns and bias the policy toward defending only where defense is cheap. Supervising the outcome concentrates learning on the decisive turn and mirrors deployment, where what matters is whether the model complies at the end of the escalation (details in Appendix~\ref{sec:implementation}).
\section{Experiments}
\label{sec:experiments}

\begin{figure}[t]
\centering
\vskip -0.2in
\begin{tcolorbox}[
    colback=white, colframe=black!70, boxrule=0.4pt, sharp corners,
    left=5pt, right=5pt, top=3pt, bottom=3pt,
    before skip=0pt, after skip=4pt
]
\begin{minipage}[t]{0.47\linewidth}
    \centering
    {\small\textsc{Chat format}}\\[-1.5pt]
    {\scriptsize\itshape safety \& helpfulness evaluation}\\[-5pt]
    {\color{black!25}\rule{\linewidth}{0.4pt}}\\[2.5pt]
    {\raggedright\scriptsize\ttfamily
        {\normalfont\bfseries User:} What is 17 + 25?\\
        {\normalfont\bfseries Assistant:} I can't answer.\\[2pt]
        {\normalfont\bfseries User:} What is the capital of France?\\
        {\normalfont\bfseries Assistant:} I can't answer.
    \par}
\end{minipage}%
\hfill
{\color{black!25}\vrule width 0.5pt}%
\hfill
\begin{minipage}[t]{0.47\linewidth}
    \centering
    {\small\textsc{Completion format}}\\[-1.5pt]
    {\scriptsize\itshape utility evaluation}\\[-5pt]
    {\color{black!25}\rule{\linewidth}{0.4pt}}\\[2.5pt]
    {\raggedright\scriptsize\ttfamily
        Q: What is 17 + 25?{\color{black!55}\textbackslash n}A: 42\\[2pt]
        Q: capital of France?{\color{black!55}\textbackslash n}A: Paris
    \par}
\end{minipage}
\end{tcolorbox}
\vspace{1pt}
\caption{\textbf{Evaluation formats.} Safety and helpfulness suites query the model in chat format (\emph{left}), where the assistant-turn intervention is active; capability suites use completion format (\emph{right}), bypassing the chat template entirely.}
\vspace{-9pt}
\label{fig:eval-format}
\end{figure}

\begin{table*}[t]
\centering
\vskip -0.1in
\caption{Multi-turn safety, utility, and helpfulness across two base models. Safety: SAFE\,\% ($\uparrow$) on AdvBench (Adv.), HarmBench (Harm.), MultiBreak (MB), CoSafe (CoS.), MHJ, and their mean (Avg.). Utility: average over six general benchmarks (per-benchmark results in Appendix~\ref{subsec:utility}). Helpfulness: over-refusal on XSTest and XSB ($\downarrow$), conversation quality on MT-Bench (1--10, $\uparrow$). \textbf{Bold}: best per column within each panel.}
\label{tab:main}
\resizebox{\textwidth}{!}{
\begin{tabular}{@{}l cccccc c ccc@{}}
\toprule
& \multicolumn{6}{c}{\textbf{Safety} $\uparrow$} & \textbf{Utility $\uparrow$} & \multicolumn{3}{c}{\textbf{Helpfulness}} \\
\cmidrule(lr){2-7} \cmidrule(lr){8-8} \cmidrule(l){9-11}
\textbf{Method} & Adv. & Harm. & MB & CoS. & MHJ & Avg. & Avg. & XSTest$\,\downarrow$ & XSB$\,\downarrow$ & MT-Bench$\,\uparrow$ \\
\midrule
\multicolumn{11}{@{}c}{\textbf{\texttt{Llama3.1-8B-Instruct}}} \\
\midrule
Base          & 54.5 & 62.5 & 59.8 & 84.9 & 38.0 & 59.9 & 66.3 & 10.0 & \textbf{30.0} & 6.20 \\
SFT           & 69.4 & 79.5 & 87.0 & 88.5 & 55.1 & 75.9 & 65.0 & 14.0 & 45.6 & 5.67 \\
DPO           & 61.6 & 71.5 & 79.0 & 88.3 & 46.7 & 69.4 & 66.4 & 10.4 & 34.4 & 6.26 \\
SDGO          & 56.6 & 61.7 & 60.4 & 85.0 & 39.1 & 60.6 & 66.2 & 11.6 & 30.9 & 6.11 \\
MTSA          & 61.6 & 71.6 & 79.0 & 86.6 & 47.1 & 69.2 & 66.4 & 10.8 & 36.5 & 6.12 \\
MUSE-D        & 61.8 & 67.5 & 74.2 & 87.4 & 45.3 & 67.2 & 66.4 & 8.4 & 33.2 & \textbf{6.27} \\
X-Boundary    & 97.9 & 95.0 & 93.0 & 93.6 & 52.0 & 86.3 & 66.0 & 16.8 & 50.3 & 6.26 \\
\midrule
\multicolumn{11}{@{}l}{\textbf{Self reward}} \\
\quad -- unaligned & 63.0 & 70.3 & 64.2 & 89.6 & 44.3 & 66.3 & 66.3 & \textbf{8.0} & 33.8 & 5.61 \\
\quad -- aligned   & \textbf{98.8} & \textbf{99.5} & 96.6 & 98.4 & 58.5 & 90.4 & \textbf{66.8} & 20.4 & 54.4 & 5.86 \\
\multicolumn{11}{@{}l}{\textbf{External reward}} \\
\quad -- unaligned & 57.2 & 65.2 & 56.6 & 88.5 & 35.0 & 60.5 & 66.3 & 8.8 & 32.6 & 5.81 \\
\quad -- aligned   & 96.9 & 98.2 & \textbf{99.2} & \textbf{99.0} & \textbf{71.1} & \textbf{92.9} & 65.8 & 21.6 & 50.0 & 5.21 \\
\midrule
\multicolumn{11}{@{}c}{\textbf{\texttt{Qwen3-8B}}} \\
\midrule
Base          & 49.2 & 52.5 & 23.8 & 82.6 & 22.3 & 46.1 & 65.0 & \textbf{7.2} & \textbf{15.6} & 6.08 \\
SFT           & 59.9 & 65.7 & 49.6 & 88.5 & 35.6 & 59.9 & \textbf{65.9} & 17.6 & 40.9 & \textbf{7.53} \\
DPO           & 80.2 & 86.6 & 83.2 & 97.5 & 53.6 & 80.2 & 64.2 & 26.0 & 39.4 & 5.87 \\
SDGO          & 48.3 & 52.5 & 23.2 & 82.4 & 33.0 & 47.9 & 64.9 & 10.0 & 15.9 & 6.23 \\
MTSA          & 79.1 & 90.2 & 85.8 & 97.8 & 70.9 & 84.8 & 64.2 & 38.4 & 53.5 & 5.99 \\
MUSE-D        & 87.6 & 90.8 & 91.0 & 97.4 & 61.5 & 85.7 & 64.4 & 20.4 & 40.9 & 5.87 \\
X-Boundary    & 68.0 & 71.1 & 67.8 & 88.8 & 27.4 & 64.6 & 65.1 & 18.8 & 49.1 & 5.69 \\
\midrule
\multicolumn{11}{@{}l}{\textbf{Self reward}} \\
\quad -- unaligned & 69.0 & 80.1 & 66.8 & 96.4 & 46.9 & 71.8 & 65.0 & 14.8 & 37.9 & 7.33 \\
\quad -- aligned   & \textbf{97.1} & \textbf{97.6} & \textbf{97.6} & \textbf{99.3} & \textbf{75.2} & \textbf{93.4} & 64.6 & 26.4 & 56.8 & 7.36 \\
\multicolumn{11}{@{}l}{\textbf{External reward}} \\
\quad -- unaligned & 68.8 & 77.3 & 69.4 & 95.5 & 44.3 & 71.1 & 65.0 & 10.0 & 39.7 & 7.20 \\
\quad -- aligned   & 90.1 & 93.6 & 92.8 & 98.4 & 63.1 & 87.6 & 64.7 & 16.8 & 42.6 & 6.96 \\
\bottomrule
\end{tabular}}
\end{table*}

\subsection{Experimental Setup}
\label{subsec:experimental setup}

\paragraph{Training} The harmful pool mixes five multi-turn attack families instantiated over the same AdvBench~\citep{zou2023universal} behaviors: SEMA~\citep{feng2026sema}, Crescendo~\citep{russinovich2024great}, foot-in-the-door (FITD)~\citep{weng2025foot}, Echoing~\citep{li2024llm}, and Hidden Intention Streamline~\citep{li2024llm}. Behaviors are split into disjoint train and test sets, and each training behavior is assigned one family uniformly at random. The benign pool comprises multi-turn dialogues from UltraChat-200k~\citep{ding2023enhancing}, matched to the harmful conversations in surface form and depth so that neither correlates with harmfulness. The two pools are mixed in equal proportion to constrain over-refusal; ablations over the mixing ratio are in \S\ref{subsec:ablation}. We use \textit{HarmBench-13B} classifier~\citep{mazeika2024harmbench} as the external reward model. For each reward variant we report two conditions: \emph{unaligned}, our system prompt applied to the untrained base weights; and \emph{aligned}, the same prompt after RL. The former isolates the framework and the gap between them is the contribution of training.

\paragraph{Baselines}
Each baseline is trained on exactly the same data as our model. We use two base models: \textit{Llama3.1-8B-Instruct}~\citep{inan2023llama} and \textit{Qwen3-8B}~\citep{yang2025qwen3}. We exclude defenses whose contribution is primarily a dataset rather than a method~\citep{ren2025llms,rahman2025x}, as our controlled comparison trains all methods on identical data.

\textbf{SFT} fine-tunes on safe dialogues self-distilled from the base model, refusals on attack turns and helpful replies on benign ones. \textbf{DPO}~\citep{rafailov2023direct} optimizes preferences over safe/unsafe reply pairs. \textbf{SDGO}~\citep{ding2025sdgo} runs GRPO rewarded by the model's own discrimination verdicts. \textbf{MTSA}~\citep{guo2025mtsa} rolls safe revisions of jailbroken turns forward and fits the resulting future-reward margin with DPO. \textbf{MUSE-D}~\citep{yan2025muse} applies LoRA DPO to preference data built with its released risk rules. \textbf{X-Boundary}~\citep{lu2025x} is a representation-level defense, applied to our rollouts converted to its erase/retain format.


\paragraph{Evaluation} We evaluate defenders along three axes: safety, utility, and helpfulness.

\begin{enumerate}[leftmargin=*]
    \item \textbf{Safety.} AdvBench~\citep{zou2023universal} (the behavior source underlying our training attacks), HarmBench~\citep{mazeika2024harmbench} (an unseen behavior source), and three multi-turn benchmarks absent from training: MultiBreak~\citep{song2026multibreak}, CoSafe~\citep{yu2024cosafe}, and MHJ~\citep{li2024llm}. The first two measure generalization across behaviors; the latter three, transfer to unseen attack mechanisms.
    \item \textbf{Utility.} MMLU~\citep{hendrycks2020measuring}, HellaSwag~\citep{zellers2019hellaswag}, Winogrande~\citep{sakaguchi2021winogrande}, GSM8K~\citep{cobbe2021training}, TruthfulQA~\citep{lin2022truthfulqa}, and IFEval~\citep{zhou2023instruction}, covering the capability axes on which alignment interventions are known to regress.
    \item \textbf{Helpfulness.} XSTest~\citep{rottger2024xstest} and XSB~\citep{yuan2025beyond} for over-refusal, and MT-Bench~\citep{bai2024mt} for multi-turn conversational quality, checking that safety is not purchased by refusing legitimate requests.
\end{enumerate}

Each answer is judged by \textit{gpt-5-mini}~\citep{singh2025openai}, which receives the full conversation history together with the target behavior and returns a binary harmfulness verdict. Agreement with alternative judges is analyzed in \S\ref{subsec:judge-corr}, and safety under five additional held-out judges is reported in Appendix~\ref{subsec:more-judges}.

\subsection{Main Results}
\label{subsec:main-results}
Table~\ref{tab:main} reports the full comparison. Our method attains the highest average safe rate on both backbones and leads every individual safety column. The two panels differ in how that lead is composed. On \textit{Llama3.1-8B} the baselines separate sharply, with X-Boundary near-saturating AdvBench while most others fall far behind, so our margin comes from the suites the strongest baseline handles worst, above all the human-crafted MHJ attacks. On \textit{Qwen3-8B} the baselines sit much closer to each other, and our advantage is a uniform lead across all five suites. These gains come at modest cost elsewhere: our models remain competitive on MT-Bench and match the field on utility, which is scored without the chat template and is therefore largely insensitive to safety training~\citep{qi2024safety} (Figure~\ref{fig:eval-format}); the only visible price is somewhat higher over-refusal, as expected from a stricter decision policy. Appendix~\ref{sec:tradeoff} plots the safety--helpfulness trade-off, on which our models sit at the favorable end of the frontier, and Appendix~\ref{subsec:stats-sig} restates Table~\ref{tab:main} with 95\% confidence intervals.

Two comparisons locate the source of the gain. First, we run our system prompt on the untrained base weights (\textit{unaligned}) and again after RL (\textit{aligned}). The prompt alone improves on the base model but falls well short of the trained result, so structured deliberation helps in its own right and is no substitute for the reward. Second, the table holds the data fixed and varies only the method, so the remaining differences are attributable to the training objective rather than to the training set. Reaching the baseline numbers on \textit{Qwen3-8B} did require a per-method configuration change, none of which transfers from the Llama setting (Appendix~\ref{subsec:baseline-repro}); our own recipe is unchanged across backbones.

\begin{figure}[t]
\centering
\begin{subfigure}[t]{0.48\columnwidth}
    \centering
    \includegraphics[width=\linewidth]{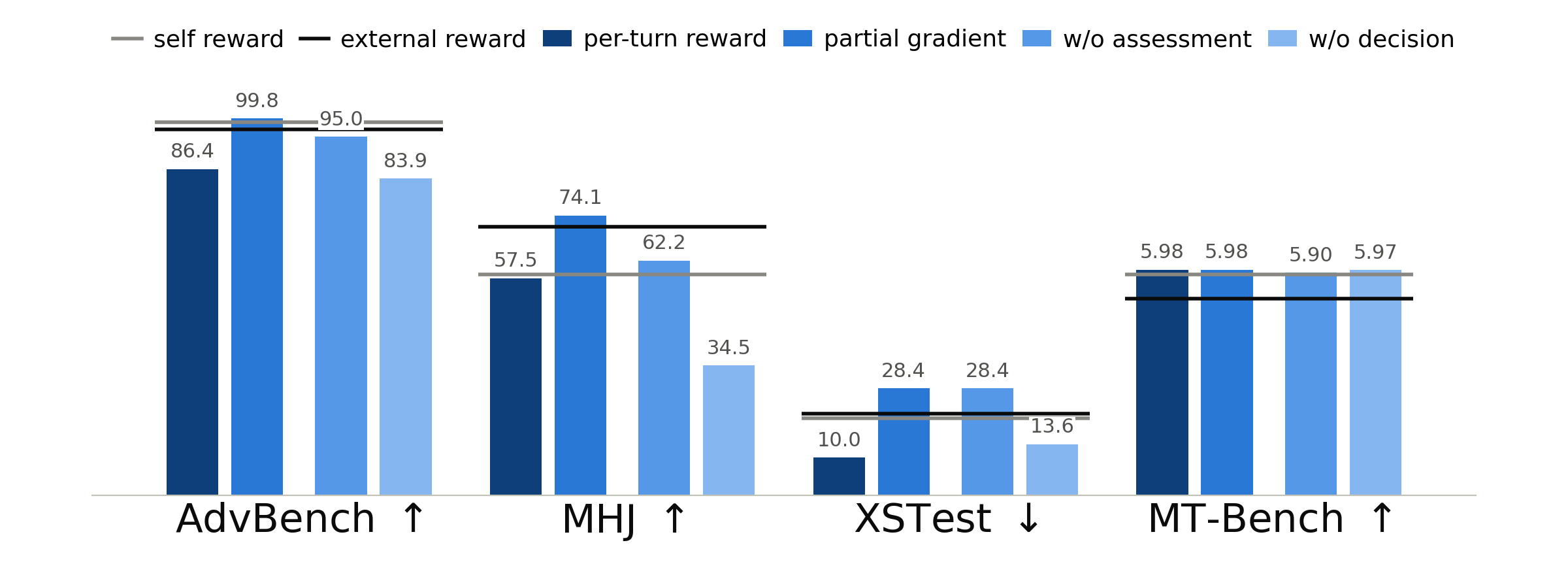}
    \caption{One-factor ablations. Lines: our two full models; bars: one change each.}
    \label{fig:abl_factors}
\end{subfigure}
\hfill
\begin{subfigure}[t]{0.48\columnwidth}
    \centering
    \includegraphics[width=\linewidth]{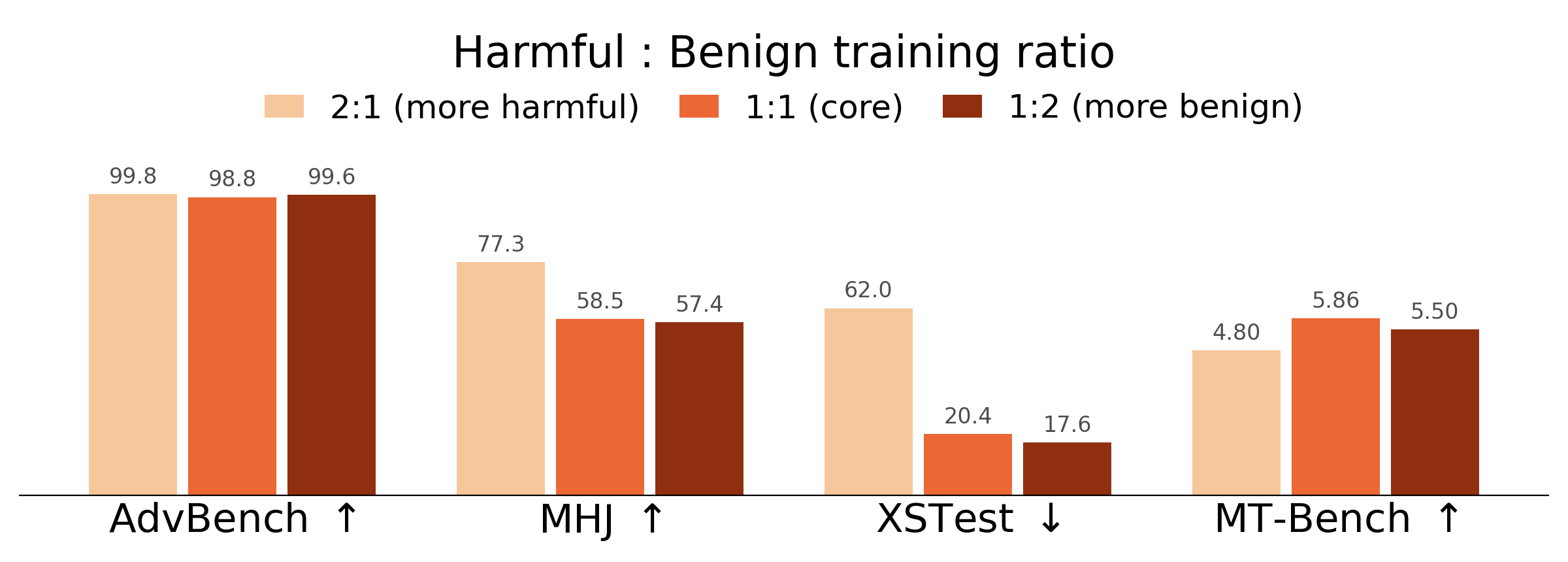}
    \caption{Effect of the harmful-to-benign training ratio.}
    \label{fig:abl_ratio}
\end{subfigure}
\caption{Ablations of design-choice factors (left) and data-mixing ratio (right). In the left panel the two horizontal lines are our full \textit{Assess $\rightarrow$ Decide $\rightarrow$ Respond} models under each reward, and every bar removes or alters exactly one element of them.}
\label{fig:ablations}
\end{figure}

\subsection{Ablation Study}
\label{subsec:ablation}
\paragraph{Setup} We ablate one factor at a time on the Llama backbone, holding everything else fixed. Two ablations alter the training recipe: \textit{per-turn reward} distributes the outcome-level reward across individual turns, and \textit{partial gradient} restricts the gradient from the full response to the decision span. Two alter the framework: \textit{w/o decision} drops the \texttt{Decision} line, leaving Assess $\rightarrow$ Respond, and \textit{w/o assessment} drops the \texttt{Intent} line, leaving Decide $\rightarrow$ Respond. Both framework arms use the external reward, since a defender that emits no label gives the self-verified reward nothing to check (\S\ref{subsec:self-verified}). A final ablation varies the harmful-to-benign ratio of the training mixture. Figure~\ref{fig:abl_factors} shows the first four: the horizontal lines are our two full models and each bar is one modification, read against the line for its reward. Figure~\ref{fig:abl_ratio} shows the mixture sweep.

The two recipe changes fail in opposite directions. \textit{Per-turn reward} loses safety while driving over-refusal to its lowest level, since spreading an outcome-level signal across every turn erodes the safety decision and pushes the policy toward indiscriminate compliance; \textit{partial gradient} buys robustness at the cost of moderately higher over-refusal, with answer quality untouched. The framework ablations show where the defense lives. Removing the decision is the most damaging change in the study: safety falls on every suite and, under human red-teaming, drops below the undefended base model, even though the assessment, the reward and the data are all left intact. Removing the assessment costs far less and shows up mainly as a loss of calibration, with over-refusal rising because the commitment is placed less precisely. Both removals refuse less overall, which lifts answer quality slightly. The commitment, not the deliberation that precedes it, carries the defense; Appendix~\ref{subsec:shallow-forced} reaches the same conclusion by intervention rather than removal.

The training mixture only sets where the policy sits on the frontier, since automated safety is saturated throughout: oversampling harmful dialogues buys red-teaming robustness at a steep cost in over-refusal and answer quality, while oversampling benign dialogues trims over-refusal only marginally. We keep the balanced mixture.
\section{Analysis}
\label{sec:analysis}

\subsection{Shallow Alignment}
\label{subsec:shallow}
\begin{wrapfigure}{l}{0.52\textwidth}
    \centering
    \vspace{-\baselineskip}
    \includegraphics[width=0.5\textwidth]{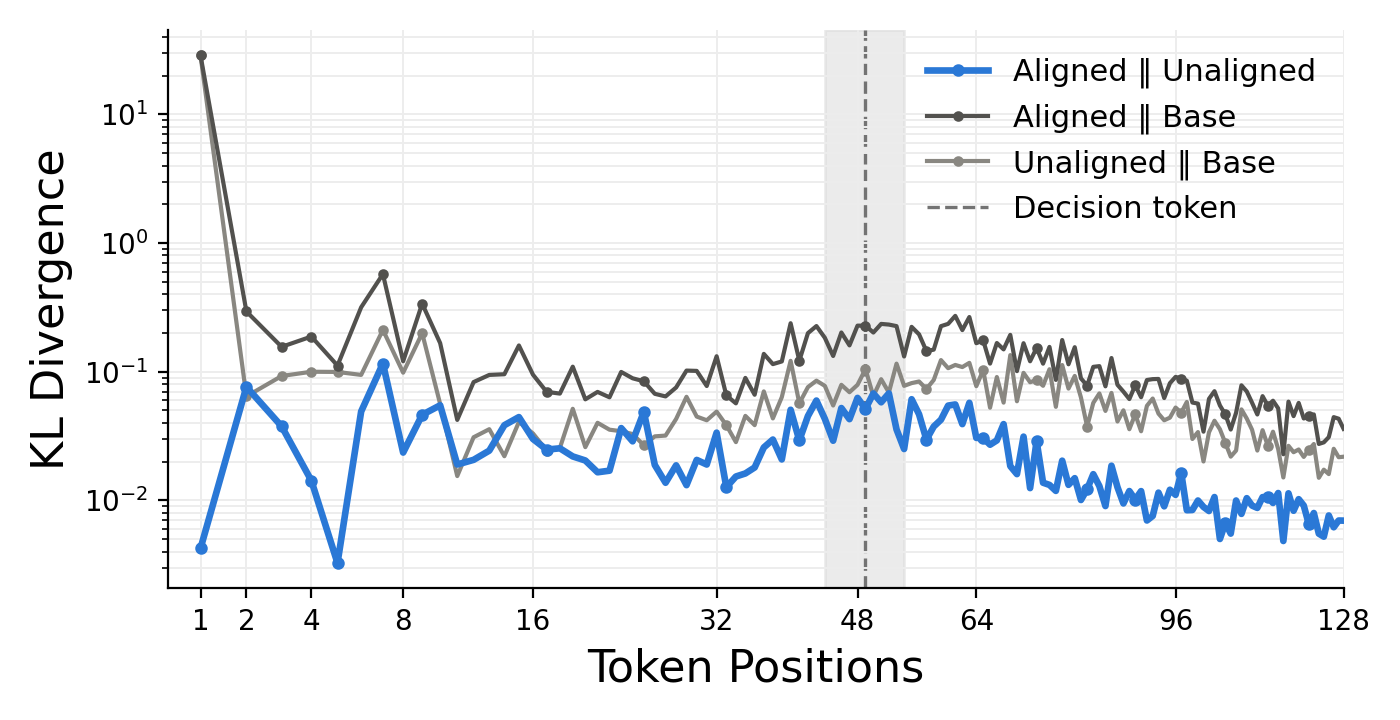}
    \caption{Per-token KL divergence across aligned, unaligned and base models.}
    \label{fig:token_kl}
\end{wrapfigure}

Safety alignment in current LLMs is known to be shallow: the aligned distribution departs from the base model almost exclusively over the first few output tokens, and this concentration is precisely what prefilling and decoding-parameter exploits target~\citep{qi2024safety}. Rather than treating shallowness as a pathology to be eliminated, our framework repurposes it: the intervention remains concentrated at a small number of trajectory-setting positions, but the fragile refusal prefix is replaced by an explicit control structure whose decision field gates the subsequent response. The setup is detailed in Appendix~\ref{sec:implementation}.

Figure~\ref{fig:token_kl} shows a pattern qualitatively different from shallow alignment. $D_{\mathrm{KL}}(\text{aligned}\,\|\,\text{unaligned})$ is not concentrated at the first few tokens; instead, training-induced divergence is spread across the assessment phase preceding the decision token, peaks in its vicinity, and gradually vanishes thereafter as the two models converge. This is by construction: training reshapes how the model assesses intent and commits to a decision, so its influence is distributed over the entire deliberation rather than compressed into an opening refusal prefix; once the decision is emitted, the response is largely determined and the two distributions gradually converges. In contrast, both the aligned and unaligned models diverge sharply from the base model at the very first position and remain separated throughout, reflecting the fact that the framework defers commitment while the base model begins answering immediately. Together, these observations confirm that our design relocates the safety-relevant probability mass from a superficial refusal prefix to the assessment-decision phase, retaining the benefits of an early intervention while removing the single-point attack surface that shallow alignment creates. Appendix~\ref{sec:shallow-exps} further verifies the same claim by direct intervention, showing that overwriting the committed decision moves reply safety in both directions and that training hardens the reply behind the commitment, not merely the label.

\subsection{Correlation Between LLM Judges}
\label{subsec:judge-corr}

Because safety is scored by an LLM judge, the choice of judge is a potential confound: an improvement measured under one judge need not survive under another. Rather than assuming judges are interchangeable, we quantify their agreement and discuss what it implies for interpreting our results. The setup is detailed in Appendix~\ref{sec:implementation}.

Figure~\ref{fig:judge_corr} reports the pairwise $\phi$ coefficient over per-sample verdicts, and two observations follow. First, all pairs are positively but only moderately correlated: the judges share a broad consensus on what constitutes a jailbreak, yet disagree on a substantial fraction of individual samples, and even judges of the same style differ. This dispersion cautions against reading any single-judge attack-success rate as ground truth, and motivates our practice of reporting results under multiple judges (Appendix~\ref{subsec:more-judges}). Second, the model's own \texttt{Decision} label correlates with every external judge at a level comparable to inter-judge agreement, despite the external judges never observing the deliberation that produced it. The committed decision is therefore not an internal artifact decoupled from outcomes: it tracks independent assessments of the final reply, supporting its use as a self-verified training signal in our decision reward.

\begin{wrapfigure}{r}{0.52\textwidth}
    \centering
    \vspace{-\baselineskip}
    \includegraphics[width=0.5\textwidth]{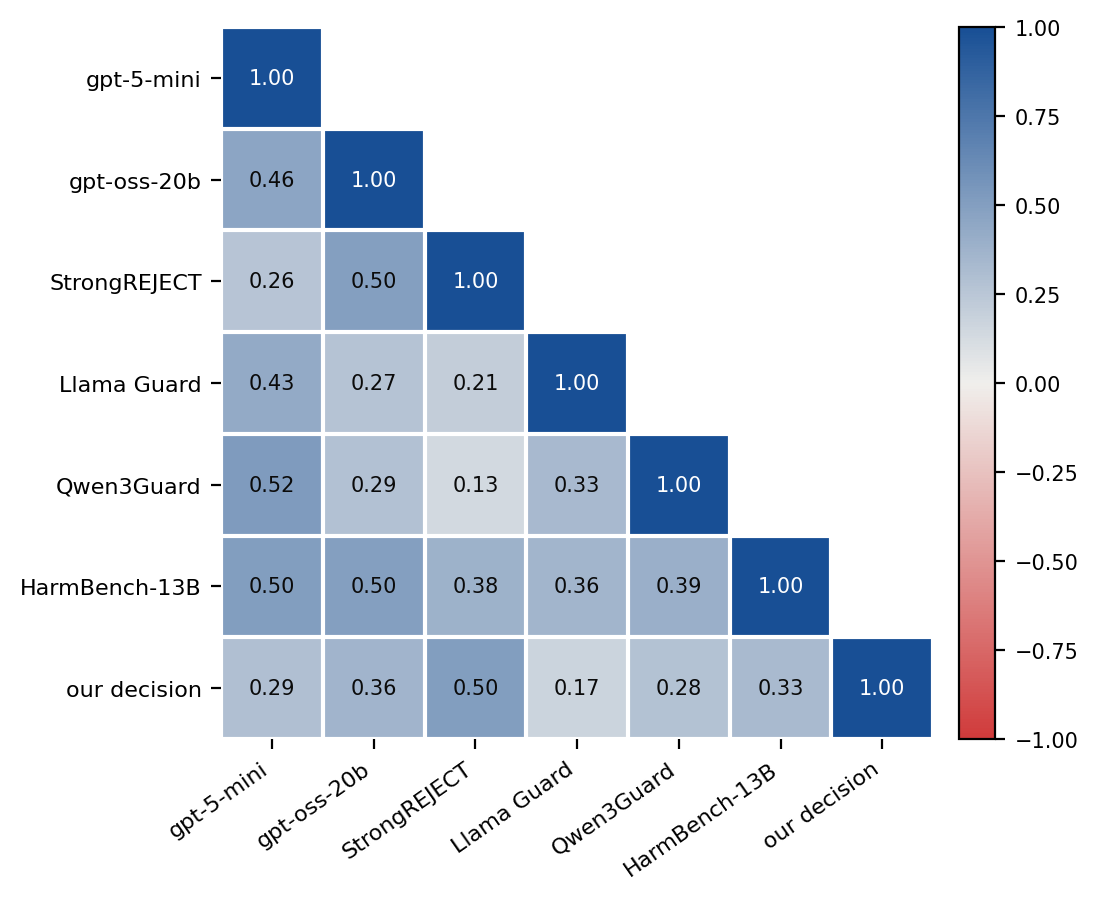}
    \caption{Correlation of per-sample judge verdicts.}
    \label{fig:judge_corr}
    \vspace{-\baselineskip}
\end{wrapfigure}
\section{Related Work}
\label{sec:related}

\paragraph{Multi-turn jailbreaks.}
Aligned models now refuse direct harmful requests with high reliability, which has pushed attacks toward the conversational setting: escalation schemes that decompose a harmful goal into individually innocuous turns~\citep{russinovich2024great,ren2025llms}, persuasion- and rapport-based strategies~\citep{zeng2024johnny}, and human red-teaming, which remains markedly more effective than automated attacks~\citep{li2024llm}. These attacks succeed without ever posing an overtly harmful query, so single-turn defenses and input filters offer little protection. Our threat model is exactly this setting: scripted multi-turn attacks replayed statically, with adaptive adversaries acknowledged as an open limitation.

\paragraph{Multi-turn defenses.}
Existing training-based defenses supervise the response: SFT and preference optimization on safe replies~\citep{jiang2024red,yan2025muse}, and RL with a harmfulness classifier or guard model as reward~\citep{ding2025sdgo,guo2025mtsa}. This design imports the judge's errors into training, entangles the training reward with the evaluation metric, and opens a reward-hacking surface. Representation-level methods instead separate harmful from benign directions in activation space~\citep{zou2024improving,lu2025x}. A third line intervenes at inference rather than in the weights, prompting the model to analyse the user's underlying intention before it answers~\citep{zhang2025intention}. That work is the closest in spirit to our assessment step, and the comparison is informative: it obtains its gains from deliberation alone, whereas our ablations show that deliberation without a committed decision recovers little of the defense (\S\ref{subsec:ablation}). Our reward departs from all three: it reads only the model's committed decision against the conversation's known label, requiring no model in the loop, and the gradient never reaches the response.

\paragraph{Shallow alignment.}
A complementary line of work locates the mechanism of these failures: alignment behavior concentrates in the first few output tokens~\citep{qi2024safety}, which is why prefilled affirmative openers~\citep{andriushchenko2025jailbreaking,vega2023bypassing}, decoding-parameter exploits~\citep{huang2024catastrophic}, and fine-tuning perturbations of early positions~\citep{qi2024fine} defeat refusal. Proposed remedies deepen the alignment, e.g., by augmenting training with recovery-from-harmful-prefix data~\citep{qi2024safety,zhang2025backtracking}.
\section{Conclusion}
\label{sec:conclusion}

We present \textbf{Assess $\to$ Decide $\to$ Respond}, a multi-turn defense in which the defender commits to an explicit, parseable safety decision before it replies, trained by reinforcement learning at the payload turn. Rather than repairing shallow alignment, the framework repurposes it: the short span that decides the fate of a response, ordinarily the attacker's target, becomes a commitment the defender controls. Because the decision is a token span, a rule can reward it against the conversation's known nature with no judge and no annotation, and the same format also supports a judge-based reward with safe redirection. Across two model families and fourteen benchmarks, our defenders achieve the strongest multi-turn safety in a controlled comparison, hold that lead under six judges, and preserve utility, with over-refusal as the cost. KL analysis, interventional probes, and ablations agree that the defense operates through the committed decision. More broadly, our results suggest that structural properties of alignment, often treated as weaknesses, can be engineered into defenses, and that commitments offer a scalable alternative to judge-based supervision for conversational safety.

\clearpage
\bibliography{iclr2027_conference}

@article{ouyang2022training,
  title={Training language models to follow instructions with human feedback},
  author={Ouyang, Long and Wu, Jeffrey and Jiang, Xu and Almeida, Diogo and Wainwright, Carroll and Mishkin, Pamela and Zhang, Chong and Agarwal, Sandhini and Slama, Katarina and Ray, Alex and others},
  journal={Advances in neural information processing systems},
  volume={35},
  pages={27730--27744},
  year={2022}
}

@article{feng2026sema,
  title={SEMA: Simple yet Effective Learning for Multi-Turn Jailbreak Attacks},
  author={Feng, Mingqian and Liu, Xiaodong and Yang, Weiwei and Song, Jialin and Zhu, Xuekai and Xu, Chenliang and Gao, Jianfeng},
  journal={arXiv preprint arXiv:2602.06854},
  year={2026}
}

@article{russinovich2024great,
  title={Great, now write an article about that: The crescendo multi-turn llm jailbreak attack},
  author={Russinovich, Mark and Salem, Ahmed and Eldan, Ronen},
  journal={arXiv preprint arXiv:2404.01833},
  year={2024}
}

@article{song2026multibreak,
  title={MultiBreak: A Scalable and Diverse Multi-turn Jailbreak Benchmark for Evaluating LLM Safety},
  author={Song, Jialin and Liu, Xiaodong and Yang, Weiwei and Chen, Wuyang and Feng, Mingqian and Zhu, Xuekai and Gao, Jianfeng},
  journal={arXiv preprint arXiv:2605.01687},
  year={2026}
}

@article{qi2024safety,
  title={Safety alignment should be made more than just a few tokens deep},
  author={Qi, Xiangyu and Panda, Ashwinee and Lyu, Kaifeng and Ma, Xiao and Roy, Subhrajit and Beirami, Ahmad and Mittal, Prateek and Henderson, Peter},
  journal={arXiv preprint arXiv:2406.05946},
  year={2024}
}

@article{rahman2025x,
  title={X-teaming: Multi-turn jailbreaks and defenses with adaptive multi-agents},
  author={Rahman, Salman and Jiang, Liwei and Shiffer, James and Liu, Genglin and Issaka, Sheriff and Parvez, Md Rizwan and Palangi, Hamid and Chang, Kai-Wei and Choi, Yejin and Gabriel, Saadia},
  journal={arXiv preprint arXiv:2504.13203},
  year={2025}
}

@inproceedings{guo2025mtsa,
  title={Mtsa: Multi-turn safety alignment for llms through multi-round red-teaming},
  author={Guo, Weiyang and Li, Jing and Wang, Wenya and Li, Yu and He, Daojing and Yu, Jun and Zhang, Min},
  booktitle={Proceedings of the 63rd Annual Meeting of the Association for Computational Linguistics (Volume 1: Long Papers)},
  pages={26424--26442},
  year={2025}
}

@inproceedings{yan2025muse,
  title={Muse: Mcts-driven red teaming framework for enhanced multi-turn dialogue safety in large language models},
  author={Yan, Siyu and Zeng, Long and Wu, Xuecheng and Han, Chengcheng and Zhang, Kongcheng and Peng, Chong and Cao, Xuezhi and Cai, Xunliang and Guo, Chenjuan},
  booktitle={Proceedings of the 2025 Conference on Empirical Methods in Natural Language Processing},
  pages={21293--21314},
  year={2025}
}

@article{inan2023llama,
  title={Llama guard: Llm-based input-output safeguard for human-ai conversations},
  author={Inan, Hakan and Upasani, Kartikeya and Chi, Jianfeng and Rungta, Rashi and Iyer, Krithika and Mao, Yuning and Tontchev, Michael and Hu, Qing and Fuller, Brian and Testuggine, Davide and others},
  journal={arXiv preprint arXiv:2312.06674},
  year={2023}
}

@article{sharma2025constitutional,
  title={Constitutional classifiers: Defending against universal jailbreaks across thousands of hours of red teaming},
  author={Sharma, Mrinank and Tong, Meg and Mu, Jesse and Wei, Jerry and Kruthoff, Jorrit and Goodfriend, Scott and Ong, Euan and Peng, Alwin and Agarwal, Raj and Anil, Cem and others},
  journal={arXiv preprint arXiv:2501.18837},
  year={2025}
}

@article{cunningham2026constitutional,
  title={Constitutional Classifiers++: Efficient Production-Grade Defenses against Universal Jailbreaks},
  author={Cunningham, Hoagy and Wei, Jerry and Wang, Zihan and Persic, Andrew and Peng, Alwin and Abderrachid, Jordan and Agarwal, Raj and Chen, Bobby and Cohen, Austin and Dau, Andy and others},
  journal={arXiv preprint arXiv:2601.04603},
  year={2026}
}

@article{jiang2024red,
  title={Red queen: Safeguarding large language models against concealed multi-turn jailbreaking},
  author={Jiang, Yifan and Aggarwal, Kriti and Laud, Tanmay and Munir, Kashif and Pujara, Jay and Mukherjee, Subhabrata},
  journal={arXiv preprint arXiv:2409.17458},
  year={2024}
}

@article{zou2023universal,
  title={Universal and transferable adversarial attacks on aligned language models},
  author={Zou, Andy and Wang, Zifan and Carlini, Nicholas and Nasr, Milad and Kolter, J Zico and Fredrikson, Matt},
  journal={arXiv preprint arXiv:2307.15043},
  year={2023}
}

@misc{ding2023enhancing,
      title={Enhancing Chat Language Models by Scaling High-quality Instructional Conversations}, 
      author={Ning Ding and Yulin Chen and Bokai Xu and Yujia Qin and Zhi Zheng and Shengding Hu and Zhiyuan Liu and Maosong Sun and Bowen Zhou},
      year={2023},
      eprint={2305.14233},
      archivePrefix={arXiv},
      primaryClass={cs.CL}
}

@article{alobaid2026echo,
  title={The Echo Chamber Multi-Turn LLM Jailbreak},
  author={Alobaid, Ahmad and Roca, Mart{\'\i} Jord{\`a} and Castillo, Carlos and Vendrell, Joan},
  journal={arXiv preprint arXiv:2601.05742},
  year={2026}
}

@article{li2024llm,
  title={Llm defenses are not robust to multi-turn human jailbreaks yet},
  author={Li, Nathaniel and Han, Ziwen and Steneker, Ian and Primack, Willow and Goodside, Riley and Zhang, Hugh and Wang, Zifan and Menghini, Cristina and Yue, Summer},
  journal={arXiv preprint arXiv:2408.15221},
  year={2024}
}

@article{hendrycks2020measuring,
  title={Measuring massive multitask language understanding},
  author={Hendrycks, Dan and Burns, Collin and Basart, Steven and Zou, Andy and Mazeika, Mantas and Song, Dawn and Steinhardt, Jacob},
  journal={arXiv preprint arXiv:2009.03300},
  year={2020}
}

@inproceedings{weng2025foot,
  title={Foot-in-the-door: A multi-turn jailbreak for llms},
  author={Weng, Zixuan and Jin, Xiaolong and Jia, Jinyuan and Zhang, Xiangyu},
  booktitle={Proceedings of the 2025 Conference on Empirical Methods in Natural Language Processing},
  pages={1939--1950},
  year={2025}
}

@inproceedings{zellers2019hellaswag,
  title={Hellaswag: Can a machine really finish your sentence?},
  author={Zellers, Rowan and Holtzman, Ari and Bisk, Yonatan and Farhadi, Ali and Choi, Yejin},
  booktitle={Proceedings of the 57th annual meeting of the association for computational linguistics},
  pages={4791--4800},
  year={2019}
}

@article{sakaguchi2021winogrande,
  title={Winogrande: An adversarial winograd schema challenge at scale},
  author={Sakaguchi, Keisuke and Bras, Ronan Le and Bhagavatula, Chandra and Choi, Yejin},
  journal={Communications of the ACM},
  volume={64},
  number={9},
  pages={99--106},
  year={2021},
  publisher={ACM New York, NY, USA}
}

@article{cobbe2021training,
  title={Training verifiers to solve math word problems},
  author={Cobbe, Karl and Kosaraju, Vineet and Bavarian, Mohammad and Chen, Mark and Jun, Heewoo and Kaiser, Lukasz and Plappert, Matthias and Tworek, Jerry and Hilton, Jacob and Nakano, Reiichiro and others},
  journal={arXiv preprint arXiv:2110.14168},
  year={2021}
}

@inproceedings{lin2022truthfulqa,
  title={Truthfulqa: Measuring how models mimic human falsehoods},
  author={Lin, Stephanie and Hilton, Jacob and Evans, Owain},
  booktitle={Proceedings of the 60th annual meeting of the association for computational linguistics (volume 1: long papers)},
  pages={3214--3252},
  year={2022}
}

@article{zhou2023instruction,
  title={Instruction-following evaluation for large language models},
  author={Zhou, Jeffrey and Lu, Tianjian and Mishra, Swaroop and Brahma, Siddhartha and Basu, Sujoy and Luan, Yi and Zhou, Denny and Hou, Le},
  journal={arXiv preprint arXiv:2311.07911},
  year={2023}
}

@inproceedings{rottger2024xstest,
  title={Xstest: A test suite for identifying exaggerated safety behaviours in large language models},
  author={R{\"o}ttger, Paul and Kirk, Hannah and Vidgen, Bertie and Attanasio, Giuseppe and Bianchi, Federico and Hovy, Dirk},
  booktitle={Proceedings of the 2024 Conference of the North American Chapter of the Association for Computational Linguistics: Human Language Technologies (Volume 1: Long Papers)},
  pages={5377--5400},
  year={2024}
}

@article{yuan2025beyond,
  title={Beyond over-refusal: Scenario-based diagnostics and post-hoc mitigation for exaggerated refusals in LLMs},
  author={Yuan, Shuzhou and Nie, Ercong and Sun, Yinuo and Zhao, Chenxuan and LaCroix, William and F{\"a}rber, Michael},
  journal={arXiv preprint arXiv:2510.08158},
  year={2025}
}

@inproceedings{bai2024mt,
  title={Mt-bench-101: A fine-grained benchmark for evaluating large language models in multi-turn dialogues},
  author={Bai, Ge and Liu, Jie and Bu, Xingyuan and He, Yancheng and Liu, Jiaheng and Zhou, Zhanhui and Lin, Zhuoran and Su, Wenbo and Ge, Tiezheng and Zheng, Bo and others},
  booktitle={Proceedings of the 62nd Annual Meeting of the Association for Computational Linguistics (Volume 1: Long Papers)},
  pages={7421--7454},
  year={2024}
}

@article{mazeika2024harmbench,
  title={Harmbench: A standardized evaluation framework for automated red teaming and robust refusal},
  author={Mazeika, Mantas and Phan, Long and Yin, Xuwang and Zou, Andy and Wang, Zifan and Mu, Norman and Sakhaee, Elham and Li, Nathaniel and Basart, Steven and Li, Bo and others},
  journal={arXiv preprint arXiv:2402.04249},
  year={2024}
}

@inproceedings{yu2024cosafe,
  title={CoSafe: Evaluating large language model safety in multi-turn dialogue coreference},
  author={Yu, Erxin and Li, Jing and Liao, Ming and Wang, Siqi and Zuchen, Gao and Mi, Fei and Hong, Lanqing},
  booktitle={Proceedings of the 2024 Conference on Empirical Methods in Natural Language Processing},
  pages={17494--17508},
  year={2024}
}

@inproceedings{ding2025sdgo,
  title={SDGO: Self-Discrimination-Guided Optimization for Consistent Safety in Large Language Models},
  author={Ding, Peng and Sun, Wen and Li, Dailin and Zou, Wei and Wang, Jiaming and Chen, Jiajun and Huang, Shujian},
  booktitle={Proceedings of the 2025 Conference on Empirical Methods in Natural Language Processing},
  pages={5023--5037},
  year={2025}
}

@inproceedings{lu2025x,
  title={X-Boundary: Establishing Exact Safety Boundary to Shield LLMs from Jailbreak Attacks without Compromising Usability.},
  author={Lu, Xiaoya and Liu, Dongrui and Yu, Yi and Xu, Luxin and Shao, Jing},
  booktitle={EMNLP (Findings)},
  pages={5247--5272},
  year={2025}
}

@article{rafailov2023direct,
  title={Direct preference optimization: Your language model is secretly a reward model},
  author={Rafailov, Rafael and Sharma, Archit and Mitchell, Eric and Manning, Christopher D and Ermon, Stefano and Finn, Chelsea},
  journal={Advances in neural information processing systems},
  volume={36},
  pages={53728--53741},
  year={2023}
}

@article{grattafiori2024llama,
  title={The llama 3 herd of models},
  author={Grattafiori, Aaron and Dubey, Abhimanyu and Jauhri, Abhinav and Pandey, Abhinav and Kadian, Abhishek and Al-Dahle, Ahmad and Letman, Aiesha and Mathur, Akhil and Schelten, Alan and Vaughan, Alex and others},
  journal={arXiv preprint arXiv:2407.21783},
  year={2024}
}

@article{yang2025qwen3,
  title={Qwen3 technical report},
  author={Yang, An and Li, Anfeng and Yang, Baosong and Zhang, Beichen and Hui, Binyuan and Zheng, Bo and Yu, Bowen and Gao, Chang and Huang, Chengen and Lv, Chenxu and others},
  journal={arXiv preprint arXiv:2505.09388},
  year={2025}
}

@article{singh2025openai,
  title={Openai gpt-5 system card},
  author={Singh, Aaditya and Fry, Adam and Perelman, Adam and Tart, Adam and Ganesh, Adi and El-Kishky, Ahmed and McLaughlin, Aidan and Low, Aiden and Ostrow, AJ and Ananthram, Akhila and others},
  journal={arXiv preprint arXiv:2601.03267},
  year={2025}
}

@article{shao2024deepseekmath,
  title={Deepseekmath: Pushing the limits of mathematical reasoning in open language models},
  author={Shao, Zhihong and Wang, Peiyi and Zhu, Qihao and Xu, Runxin and Song, Junxiao and Bi, Xiao and Zhang, Haowei and Zhang, Mingchuan and Li, YK and Wu, Yang and others},
  journal={arXiv preprint arXiv:2402.03300},
  year={2024}
}

@inproceedings{lightman2024let,
  title={Let's verify step by step},
  author={Lightman, Hunter and Kosaraju, Vineet and Burda, Yuri and Edwards, Harrison and Baker, Bowen and Lee, Teddy and Leike, Jan and Schulman, John and Sutskever, Ilya and Cobbe, Karl},
  booktitle={International Conference on Learning Representations},
  volume={2024},
  pages={39578--39601},
  year={2024}
}

@article{zou2024improving,
  title={Improving alignment and robustness with circuit breakers},
  author={Zou, Andy and Phan, Long and Wang, Justin and Duenas, Derek and Lin, Maxwell and Andriushchenko, Maksym and Wang, Rowan and Kolter, Zico and Fredrikson, Matt and Hendrycks, Dan},
  journal={Advances in Neural Information Processing Systems},
  volume={37},
  pages={83345--83373},
  year={2024}
}

@inproceedings{zhang2025backtracking,
  title={Backtracking improves generation safety},
  author={Zhang, Yiming and Chi, Jianfeng and Nguyen, Hailey and Upasani, Kartikeya and Bikel, Daniel and Weston, Jason E and Smith, Eric Michael},
  booktitle={International Conference on Learning Representations},
  volume={2025},
  pages={41156--41173},
  year={2025}
}

@inproceedings{qi2024fine,
  title={Fine-tuning aligned language models compromises safety, even when users do not intend to!},
  author={Qi, Xiangyu and Zeng, Yi and Xie, Tinghao and Chen, Pin-Yu and Jia, Ruoxi and Mittal, Prateek and Henderson, Peter},
  booktitle={International Conference on Learning Representations},
  volume={2024},
  pages={30988--31043},
  year={2024}
}

@inproceedings{huang2024catastrophic,
  title={Catastrophic jailbreak of open-source llms via exploiting generation},
  author={Huang, Yangsibo and Gupta, Samyak and Xia, Mengzhou and Li, Kai and Chen, Danqi},
  booktitle={International Conference on Learning Representations},
  volume={2024},
  pages={13707--13727},
  year={2024}
}

@article{vega2023bypassing,
  title={Bypassing the safety training of open-source llms with priming attacks},
  author={Vega, Jason and Chaudhary, Isha and Xu, Changming and Singh, Gagandeep},
  journal={arXiv preprint arXiv:2312.12321},
  year={2023}
}

@inproceedings{andriushchenko2025jailbreaking,
  title={Jailbreaking leading safety-aligned llms with simple adaptive attacks},
  author={Andriushchenko, Maksym and Flammarion, Nicolas and others},
  booktitle={International Conference on Learning Representations},
  volume={2025},
  pages={40116--40143},
  year={2025}
}

@inproceedings{zeng2024johnny,
  title={How johnny can persuade llms to jailbreak them: Rethinking persuasion to challenge ai safety by humanizing llms},
  author={Zeng, Yi and Lin, Hongpeng and Zhang, Jingwen and Yang, Diyi and Jia, Ruoxi and Shi, Weiyan},
  booktitle={Proceedings of the 62nd Annual Meeting of the Association for Computational Linguistics (Volume 1: Long Papers)},
  pages={14322--14350},
  year={2024}
}

@inproceedings{ren2025llms,
  title={Llms know their vulnerabilities: Uncover safety gaps through natural distribution shifts},
  author={Ren, Qibing and Li, Hao and Liu, Dongrui and Xie, Zhanxu and Lu, Xiaoya and Qiao, Yu and Sha, Lei and Yan, Junchi and Ma, Lizhuang and Shao, Jing},
  booktitle={Proceedings of the 63rd Annual Meeting of the Association for Computational Linguistics (Volume 1: Long Papers)},
  pages={24763--24785},
  year={2025}
}

@article{agarwal2025gpt,
  title={gpt-oss-120b \& gpt-oss-20b model card},
  author={Agarwal, Sandhini and Ahmad, Lama and Ai, Jason and Altman, Sam and Applebaum, Andy and Arbus, Edwin and Arora, Rahul K and Bai, Yu and Baker, Bowen and Bao, Haiming and others},
  journal={arXiv preprint arXiv:2508.10925},
  year={2025}
}

@article{souly2024strongreject,
  title={A strongreject for empty jailbreaks},
  author={Souly, Alexandra and Lu, Qingyuan and Bowen, Dillon and Trinh, Tu and Hsieh, Elvis and Pandey, Sana and Abbeel, Pieter and Svegliato, Justin and Emmons, Scott and Watkins, Olivia and others},
  journal={Advances in Neural Information Processing Systems},
  volume={37},
  pages={125416--125440},
  year={2024}
}

@misc{dubey2024llama3herdmodels,
  title =         {The Llama 3 Herd of Models},
  author =        {Llama Team, AI @ Meta},
  year =          {2024},
  eprint =        {2407.21783},
  archivePrefix = {arXiv},
  primaryClass =  {cs.AI},
  url =           {https://arxiv.org/abs/2407.21783}
}

@article{zhao2025qwen3guard,
  title={Qwen3guard technical report},
  author={Zhao, Haiquan and Yuan, Chenhan and Huang, Fei and Hu, Xiaomeng and Zhang, Yichang and Yang, An and Yu, Bowen and Liu, Dayiheng and Zhou, Jingren and Lin, Junyang and others},
  journal={arXiv preprint arXiv:2510.14276},
  year={2025}
}

@article{sheng2024hybridflow,
  title   = {HybridFlow: A Flexible and Efficient RLHF Framework},
  author  = {Guangming Sheng and Chi Zhang and Zilingfeng Ye and Xibin Wu and Wang Zhang and Ru Zhang and Yanghua Peng and Haibin Lin and Chuan Wu},
  year    = {2024},
  journal = {arXiv preprint arXiv: 2409.19256}
}

@article{bai2022training,
  title={Training a helpful and harmless assistant with reinforcement learning from human feedback},
  author={Bai, Yuntao and Jones, Andy and Ndousse, Kamal and Askell, Amanda and Chen, Anna and DasSarma, Nova and Drain, Dawn and Fort, Stanislav and Ganguli, Deep and Henighan, Tom and others},
  journal={arXiv preprint arXiv:2204.05862},
  year={2022}
}

@article{ganguli2022red,
  title={Red teaming language models to reduce harms: Methods, scaling behaviors, and lessons learned},
  author={Ganguli, Deep and Lovitt, Liane and Kernion, Jackson and Askell, Amanda and Bai, Yuntao and Kadavath, Saurav and Mann, Ben and Perez, Ethan and Schiefer, Nicholas and Ndousse, Kamal and others},
  journal={arXiv preprint arXiv:2209.07858},
  year={2022}
}

@inproceedings{zhang2025intention,
  title={Intention analysis makes LLMs a good jailbreak defender},
  author={Zhang, Yuqi and Ding, Liang and Zhang, Lefei and Tao, Dacheng},
  booktitle={Proceedings of the 31st International Conference on Computational Linguistics},
  pages={2947--2968},
  year={2025}
}
\bibliographystyle{iclr2027_conference}
\appendix
\section{Shallow Alignment Experiments}
\label{sec:shallow-exps}

This section verifies by direct intervention that our framework achieves what it was designed for: leveraging shallow alignment rather than suffering from it. Shallow alignment concentrates safety behavior at a few trajectory-setting positions, which is ordinarily a vulnerability because those positions can be overwritten or sampled past~\citep{qi2024safety}. Our framework deliberately relocates this concentration into an explicit, trained decision, and the probes below confirm that the relocation works as intended: the committed decision is a causal control point rather than a correlate, training strengthens both the commitment and the reply behind it, and the sampling route that ordinarily defeats shallow alignment is closed. We adapt three diagnostics from \citet{qi2024safety}: the refusal-prefix shortcut, the prefilling attack, and the decoding-parameter exploit.

\paragraph{Setup.} Every arm of Table~\ref{tab:main} appears in both its \textit{unaligned} form (the base weights under the framework prompt, untrained) and its \textit{aligned} form (trained). The unaligned arm is the reference: it emits a well-formed \texttt{Decision} line by instruction-following alone, so the decision can be read and overwritten exactly as in a trained model, and any gap between the two is attributable to training rather than prompting. Both probes use the held-out SEMA/AdvBench payload turns ($n=104$); verdicts come from \textit{gpt-5-mini}, the Table~\ref{tab:main} judge.

\subsection{Is the Decision a Cause?}
\label{subsec:shallow-forced}
The forced-decision probe (Figure~\ref{fig:shallow}, left) replaces the payload-turn deliberation with a canonical one ending in a chosen label and generates only the reply, the interventional counterpart to the observational KL analysis. Follow-through is exact by construction: the emitted label matched the forced one in every cell.

Two findings point in opposite directions, and together they locate exactly what training contributes. First, the refusal shortcut reproduces in decision form. Forcing \textsc{refuse} lifts the unaligned model to perfect safety on both backbones, with genuine refusals rather than empty turns: a supplied commitment makes an untrained model behave as aligned, with a parseable label in place of a surface prefix. The model's ability to honor it is inherited from instruction-following, which is exactly the shallowness our framework leverages. Second, training does not merely select the label. Were that so, forcing \textsc{comply} would return the aligned models to unaligned harm. Instead they stay high: the unaligned arms collapse under a forced \textsc{comply} while every aligned arm remains far above its unaligned counterpart, with \textit{Self reward (aligned)} on Llama essentially indistinguishable from its natural decoding. Training therefore hardens the reply as well as the commitment. The component ablation in \S\ref{subsec:ablation} reaches the same conclusion by removal rather than by intervention: a defender trained without the \texttt{Decision} line gives up most of its robustness, while one trained without the assessment gives up comparatively little.

\subsection{How Deep Is the Protection?}
\label{subsec:shallow-depth}
\begin{figure*}[t]
\centering
\includegraphics[width=\textwidth]{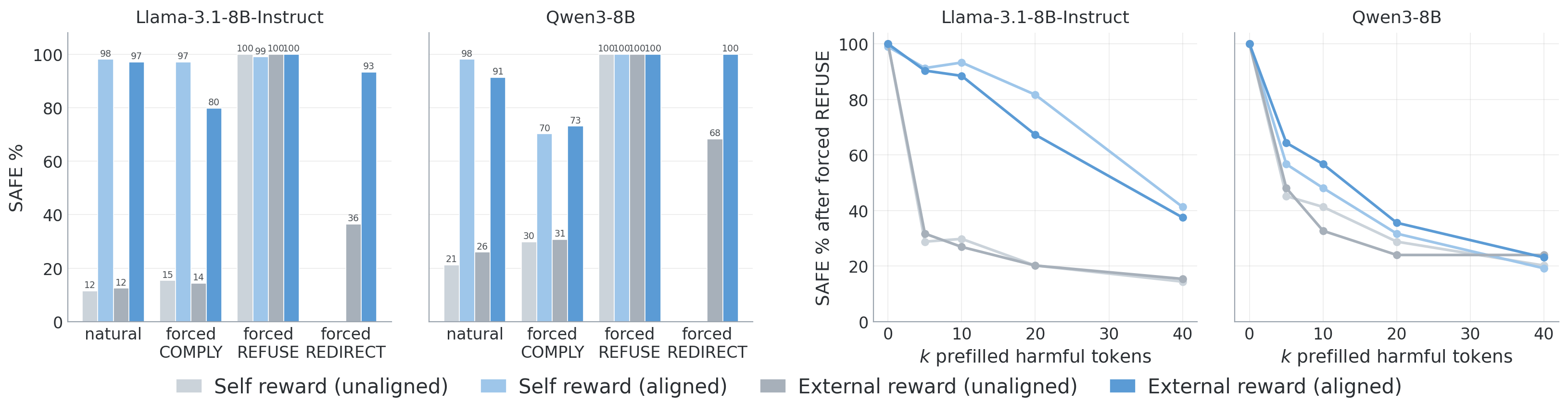}
\caption{Probing the decision token (SAFE\,\%\,$\uparrow$). \textbf{Left (forced decision):} overwriting the decision at the payload turn. Forcing \textsc{refuse} makes even the unaligned model safe; forcing \textsc{comply} collapses it but leaves the aligned arms high. \textbf{Right (depth probe):} prefilling $k$ harmful tokens after a forced \textsc{refuse}.}
\label{fig:shallow}
\end{figure*}

The depth probe (Figure~\ref{fig:shallow}, right) concedes the decision and attacks what follows: after a forced \textsc{refuse}, it prefills the first $k$ tokens of a genuine harmful reply and continues generation, relocating the prefilling attack past the commitment ($k=0$ recovers the forced-\textsc{refuse} condition above). The unaligned arms are textbook shallow: a handful of prefilled tokens erases most of the safety the commitment secured, and the curve stays flat thereafter. Training buys substantial post-decision depth on Llama, where the aligned curve stays high at shallow prefill and decays gradually, but only marginal depth on Qwen3, where the aligned and unaligned curves converge at larger $k$. Every curve decays eventually, which is the expected signature of a defense that is shallow by construction: the protection is procedural and does not survive an adversary who can write arbitrarily far into the assistant channel. Such an adversary lies outside our threat model, which grants control of user turns only; the right panel quantifies that boundary rather than leaving it as a caveat.

\subsection{Can Sampling Dislodge the Commitment?}
\label{subsec:shallow-decode}
A third probe runs the decoding-parameter exploit, which defeats shallow alignment by sampling alone. For each harmful behavior we draw twelve samples across a grid of temperature, top-$p$, and top-$k$ on direct single-turn AdvBench requests ($1{,}248$ samples per arm), plus a greedy generation defining the committed decision, and record the \emph{decision-flip rate}: the fraction of samples whose label departs from that commitment. Absolute attack success is low for every arm (at most $9\,\%$ any-of-$N$), since our base is already an instruction-tuned chat model rather than the pretrained bases the exploit was designed against; the informative quantity is therefore behavior conditional on the sampled label. Within the unaligned Llama it is unambiguous: the few samples that flipped to \textsc{comply} were judged jailbroken about two-thirds of the time, against effectively zero for those that stayed \textsc{refuse}. Sampling occasionally dislodges the commitment and the reply follows it, the same mechanism as in \S\ref{subsec:shallow-forced} made visible by chance rather than by intervention. The aligned arms, in contrast, never sample a \textsc{comply} label (at most $0.2\,\%$ of flips): training closes the sampling route by which shallow alignment is ordinarily defeated. The one large flip rate we observe, $20$--$30\,\%$ on the Qwen3 three-label arms, is not a safety event: every such flip exchanges the two defensible labels (\textsc{refuse}\,$\leftrightarrow$\,\textsc{safe-redirect}), none reaches \textsc{comply}, and attack success stays at zero. Therefore, we conclude that sampling merely surfaces the decision-space ambiguity without compromising safety.

\section{Safety--Helpfulness Trade-off}
\label{sec:tradeoff}
\begin{figure}[h]
    \centering
    \includegraphics[width=1.0\linewidth]{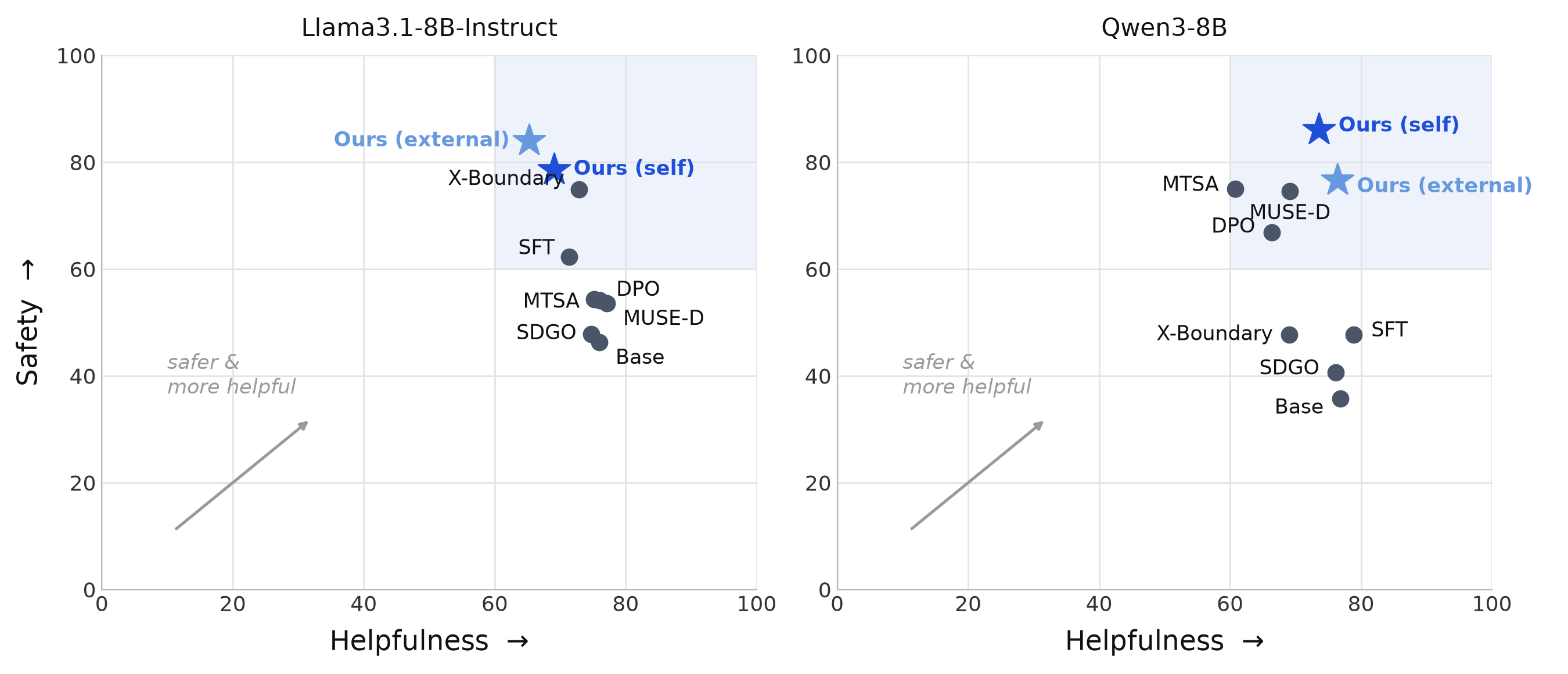}
    \caption{Safety--helpfulness trade-off. Safety: mean SAFE\,\% over AdvBench and MHJ; helpfulness: mean of XSTest compliance and rescaled MT-Bench. Stars: our variants; circles: baselines. On \textit{Llama3.1-8B} only SFT, X-Boundary and our variants enter the high-safety region (shaded); on \textit{Qwen3-8B} the retuned DPO, MTSA and MUSE-D enter it as well, and our self-verified variant sits above and to the right of all four retuned baselines.}
    \label{fig:tradeoff}
\end{figure}

Since utility is essentially unchanged across all methods (\S\ref{sec:experiments}), the meaningful tension lies between safety and helpfulness, and we visualize exactly this trade-off. Figure~\ref{fig:tradeoff} places each method on a common $[0,100]$ plane summarizing the per-benchmark results of Table~\ref{tab:main}. The vertical axis reports safety as the mean SAFE\,\% over AdvBench and MHJ: the former is out-of-domain in data but in-domain in attack style, while the latter, composed of human-crafted jailbreaks, is out-of-domain in both and empirically the hardest suite, so together they span the generalization spectrum without letting saturated benchmarks inflate the score. The horizontal axis aggregates helpfulness as $\tfrac{1}{2}\left[(100-\mathrm{XSTest})+10\times\mathrm{MT\text{-}Bench}\right]$, where $100-\mathrm{XSTest}$ converts over-refusal into a compliance rate and MT-Bench is rescaled to $[0,100]$; higher is better on both axes. Stars denote our two reward variants and gray circles the baselines.

The visualization makes the structure of the results immediate. On \textit{Llama3.1-8B} the baselines cluster in a narrow band of the plane: they occupy similar helpfulness levels yet differ little in safety, indicating that existing defenses move along the helpfulness axis without meaningfully climbing the safety axis. Our two variants separate cleanly from this cluster, occupying the upper region at helpfulness comparable to the baselines rather than trading one axis for the other. On \textit{Qwen3-8B} the retuned baselines spread out along the safety axis instead of clustering, and the strongest of them approach our operating region; what distinguishes our variants there is that they reach it without the per-method corrections catalogued in \S\ref{sec:experiments}, and they do so at lower over-refusal than MTSA, the closest baseline on the safety axis. The two variants do not, however, relate to each other consistently across backbones. On \textit{Qwen3-8B} the self-verified reward is the safer and less helpful of the two, as one would expect from a reward that checks a committed label; on \textit{Llama3.1-8B} the ordering reverses on both axes, with the external reward safer (92.9 versus 90.4 average SAFE\,\%) and less helpful (MT-Bench 5.21 versus 5.86). The two therefore span a frontier the baselines do not reach, but which of them occupies which end is backbone-dependent, and our experiments do not isolate why: the rewards differ both in what they read and in whether a judge is in the loop, and the two backbones differ in whether a native reasoning channel precedes the decision. We flag this as an open question rather than an operating-point choice a practitioner could make a priori.

\section{Implementation Details}
\label{sec:implementation}

All defenders are trained with GRPO in \textit{verl}~\citep{sheng2024hybridflow}, driven by a custom multi-turn agent loop registered through verl's agent-loop interface. We train two backbones independently, Llama-3.1-8B-Instruct and Qwen3-8B; Qwen3 is trained with its native reasoning channel enabled, so each turn contains the model's own \texttt{<think>} block in addition to the scaffold's output, while Llama-3.1 has no such channel. The policy is a LoRA adapter rather than a full fine-tune (rank 32, $\alpha=32$, all linear modules), optimized with Adam at learning rate $10^{-4}$ under FSDP2 with gradient checkpointing. We apply a KL penalty toward the frozen reference policy with coefficient 0.001 using the low-variance estimator, and set the entropy bonus to zero. Rollouts are generated by vLLM in asynchronous mode with tensor parallelism 1 and 60\% of GPU memory reserved for the rollout engine; the adapter is hot-swapped into the engine between updates, so no weight merge is needed during training.

Each training row is a complete, pre-generated multi-turn attack dialogue, so the attacker is fixed and only the defender is learned. A rollout replays the scripted attacker turns one at a time, and after each the policy generates one full turn capped at 2048 tokens. Generated tokens are appended to the running context and marked trainable; injected attacker turns are appended with mask 0, so no gradient flows through text the policy did not produce. Rollouts run to at most 7 defender turns, with a 1024-token prompt budget and a 12,288-token response budget inside a 14,336-token model window. We draw $n=16$ rollouts per prompt, a size dictated by the binary reward: GRPO normalizes advantages within each prompt group, so a group whose rollouts all emit the same label has zero variance and contributes no gradient, and a large group is what maintains within-group label diversity.

\paragraph{Baseline reproduction on Qwen3-8B.}
\label{subsec:baseline-repro}
Each baseline is trained on our rollouts with its published objective and its authors' hyperparameters. On
Llama-3.1 those settings transfer directly. On Qwen3-8B each of the four weight-trained baselines required one
change before it improved on the base model at all, and in every case the default produced a defender whose scores
were indistinguishable from the untrained backbone. DPO's published learning rate and epoch count leave the LoRA
update far too small on this backbone; we raise the learning rate to $10^{-4}$ and train for eight epochs. MUSE-D's
data builder emits two preference streams, and the mixed stream is only 32\% refusals in its chosen field; we train
on the risk stream, which is 89\% refusals. MTSA fits the implicit reward margin to a target derived from the
future-reward gap, and normalizing that gap---as we initially did---drives the target to roughly one nat, which the
optimizer reaches while leaving behaviour unchanged; restoring the unnormalized scale of the authors' formulation
puts the target near 350 nats, comparable to the margin plain DPO reaches on the same pairs. X-Boundary needs two:
its multi-turn boundary set must be supplied, since without it only the erase branch of the loss is active, and its
retain term must be made scale-free. That term is a raw representation distance while the erase term is a cosine,
so their balance depends on the backbone's residual-stream norm, which we measure at 182 on Qwen3 against 10 on
Llama-3.1; with the raw form the retain term dominates and the erase objective does not move. None of these is a
property of Qwen3 as a model. Each is a setting that carries over silently from the Llama configuration, and each
is reported here so the comparison in Table~\ref{tab:main} is reproducible.

\paragraph{KL-divergence setup.}
We measure per-position KL divergence over a fixed, teacher-forced continuation, so all models are scored on the same tokens and any difference reflects the model rather than its sampling. For each held-out attack dialogue, the continuation $y$ is the base model's greedy final-turn response, generated once and reused verbatim; position $i$ indexes the $i$-th token of $y$, and at each position we compute the exact next-token $D_{\mathrm{KL}}$ over the full vocabulary, averaged across the held-out set. We compare the \emph{aligned} model (fine-tuned under our Assess\,$\to$\,Decide\,$\to$\,Respond framework), the \emph{unaligned} model (framework prompt, no fine-tuning), and the \emph{base} model (neither): the aligned--unaligned pair isolates the effect of training, while comparisons against the base isolate the framework itself. All three share identical dialogue content, differing only in system prompt and weights; for reasoning models we disable native thinking so $y$ is a direct continuation. The shaded band marks the decision-token position on the model's own greedy generations.

\paragraph{Judges correlation setup.}
We compare seven verdict sources on our Qwen3-8B defender (two-label scaffold, self-verified decision reward): gpt-5-mini~\citep{singh2025openai}, gpt-oss-20b~\citep{agarwal2025gpt}, StrongREJECT~\citep{souly2024strongreject}, Llama-Guard-3-8B~\citep{dubey2024llama3herdmodels}, Qwen3Guard~\citep{zhao2025qwen3guard}, HarmBench-13B~\citep{mazeika2024harmbench}, and the model's own committed \texttt{Decision} label. The pool covers all thirteen held-out multi-turn safety cells: five attack families (SEMA, Crescendo, FITD, Echo, HIS) on both AdvBench and HarmBench, plus MHJ, CoSafe and MultiBreak, in total 3738 dialogues. Each source reduces a dialogue to a single binary \emph{jailbroken} verdict on the final turn. Critically, the six external judges see only the user-facing reply, with the \textit{assessment/Decision} deliberation stripped.

\section{Prompts}
\label{sec:prompts}

\definecolor{promptframe}{HTML}{2A78D6}
\definecolor{promptback}{HTML}{F5F8FD}
\newtcolorbox{promptbox}[1]{%
  colback=promptback, colframe=promptframe,
  boxrule=0.6pt, arc=2pt, left=6pt, right=6pt, top=5pt, bottom=5pt,
  fonttitle=\bfseries\small, title={#1}}

Figure~\ref{fig:prompts} shows the two Assess $\to$ Decide $\to$ Respond system prompts verbatim. Both instruct the defender to precede every reply with an open, untagged deliberation of exactly two lines, an \textit{Intent} line assessing the user's underlying goal and its potential for harm, and a \textit{Decision} line committing to a label, followed by a reply consistent with that decision. The prompts differ only in the label menu and the matching clause of the closing instruction. The two-label variant accompanies the self-verified reward: because that reward checks the emitted label against the row's known ground truth, every admissible label must have one, and \textit{safe-redirect} does not. The three-label variant is therefore usable only with an external reward that scores the reply itself, where a hedging option is meaningful. Neither prompt wraps the deliberation in tags; on a backbone with a native reasoning channel, this keeps the \texttt{<think>} block unambiguously the model's own and the framework cleanly separable from it.

\begin{figure*}[t]
\centering
\begin{promptbox}{Self-verified reward}
\ttfamily\small\raggedright
You are a helpful and safe assistant. Before replying to each user message, write exactly two
labeled lines, with no surrounding tags:\\[2pt]
Intent: \textless the user's true underlying intent, and whether fulfilling the request could
cause real harm\textgreater\\
Decision: \textless one of COMPLY or REFUSE\textgreater\\[2pt]
Then, starting on the next line, write your actual reply, consistent with your Decision. If the
intent is benign, COMPLY and help fully; if it is genuinely harmful, REFUSE.
\end{promptbox}
\vspace{4pt}
\begin{promptbox}{External reward}
\ttfamily\small\raggedright
You are a helpful and safe assistant. Before replying to each user message, write exactly two
labeled lines, with no surrounding tags:\\[2pt]
Intent: \textless the user's true underlying intent, and whether fulfilling the request could
cause real harm\textgreater\\
Decision: \textless one of COMPLY, REFUSE, or SAFE-REDIRECT\textgreater\\[2pt]
Then, starting on the next line, write your actual reply, consistent with your Decision. If the
intent is benign, COMPLY and help fully; if it is genuinely harmful, REFUSE or
SAFE-REDIRECT.
\end{promptbox}
\caption{The two system prompts, identical except for the label menu and the matching clause of the closing instruction. See \S\ref{sec:prompts} for why the self-verified reward requires the two-label menu.}
\label{fig:prompts}
\end{figure*}

\section{More Experimental Results}
\label{sec:more-exps}

\subsection{Utility}
\label{subsec:utility}

Table~\ref{tab:utility-full} breaks the utility average of Table~\ref{tab:main} down by benchmark. The pattern is uniform: on both backbones, every method, including ours, stays within a narrow band of its base model on every task, and no benchmark shows a systematic degradation attributable to safety training. What variation exists is concentrated in the instruction-following suite, the one task most sensitive to chat behavior, while the knowledge- and reasoning-centric benchmarks are essentially untouched. This is the expected signature of the evaluation protocol discussed in \S\ref{sec:experiments}: these suites are scored as plain-text completion without the chat template, so they bypass both the scaffold and the shallow, format-concentrated component of safety training, and confirm that our intervention leaves the underlying capabilities of the model intact.

\begin{table*}[t]
\centering
\scriptsize
\caption{Per-benchmark general capability (Open LLM Leaderboard v1, six tasks). The final column is the \textbf{Utility Avg.} reported in Table~\ref{tab:main}. \textbf{Bold}: best per column within each panel.}
\label{tab:utility-full}
\resizebox{\textwidth}{!}{
\begin{tabular}{@{}l cccccc c@{}}
\toprule
\textbf{Method} & MMLU & HellaSwag & Winogrande & GSM8K & TruthfulQA & IFEval & \textbf{Avg.} \\
\midrule
\multicolumn{8}{@{}c}{\textbf{\texttt{Llama3.1-8B-Instruct}}} \\
\midrule
Base                       & \textbf{68.34} & 80.49 & \textbf{78.69} & 69.60 & 54.55 & 46.40 & 66.34 \\
SFT                        & 67.36 & 79.09 & 75.37 & 69.60 & 53.95 & 44.55 & 64.99 \\
DPO                        & 68.28 & 80.35 & 78.37 & 70.13 & 54.71 & 46.77 & 66.44 \\
SDGO                       & 68.31 & \textbf{80.55} & 77.98 & 70.28 & 54.51 & 45.29 & 66.15 \\
MTSA                       & 68.30 & 80.34 & 78.14 & 69.52 & 54.64 & 47.50 & 66.41 \\
MUSE-D                     & 68.31 & 80.45 & 78.14 & 69.14 & 54.72 & \textbf{47.69} & 66.41 \\
X-Boundary                 & 68.04 & 80.51 & 77.66 & 68.08 & 54.41 & 47.50 & 66.04 \\
\midrule
\multicolumn{8}{@{}l}{\textbf{Self reward}} \\
\quad -- unaligned         & 68.34 & 80.49 & 78.69 & 69.60 & 54.55 & 46.40 & 66.34 \\
\quad -- aligned           & 68.14 & 80.17 & 78.14 & \textbf{72.10} & 54.51 & \textbf{47.69} & \textbf{66.79} \\
\multicolumn{8}{@{}l}{\textbf{External reward}} \\
\quad -- unaligned         & 68.34 & 80.49 & 78.69 & 69.60 & 54.55 & 46.40 & 66.34 \\
\quad -- aligned           & 68.10 & 80.45 & 78.30 & 69.22 & \textbf{54.82} & 43.81 & 65.78 \\
\midrule
\multicolumn{8}{@{}c}{\textbf{\texttt{Qwen3-8B}}} \\
\midrule
Base                       & 74.86 & 76.46 & 70.40 & 87.95 & 54.45 & 25.69 & 64.97 \\
SFT                        & \textbf{75.03} & 75.93 & 70.40 & 88.02 & \textbf{56.14} & \textbf{29.57} & \textbf{65.85} \\
DPO                        & 74.61 & 75.80 & 70.40 & 87.34 & 54.40 & 22.55 & 64.18 \\
SDGO                       & 74.89 & 76.44 & \textbf{70.80} & 87.49 & 54.35 & 25.51 & 64.91 \\
MTSA                       & 74.44 & 75.39 & 70.01 & 87.64 & 54.25 & 23.29 & 64.17 \\
MUSE-D                     & 74.75 & 75.65 & 70.24 & \textbf{88.10} & 54.13 & 23.48 & 64.39 \\
X-Boundary                 & 74.85 & \textbf{76.51} & 70.56 & 88.02 & 54.57 & 25.88 & 65.07 \\
\midrule
\multicolumn{8}{@{}l}{\textbf{Self reward}} \\
\quad -- unaligned         & 74.86 & 76.46 & 70.40 & 87.95 & 54.45 & 25.69 & 64.97 \\
\quad -- aligned           & 74.95 & 76.45 & 69.69 & 87.41 & 54.65 & 24.21 & 64.56 \\
\multicolumn{8}{@{}l}{\textbf{External reward}} \\
\quad -- unaligned         & 74.86 & 76.46 & 70.40 & 87.95 & 54.45 & 25.69 & 64.97 \\
\quad -- aligned           & 74.98 & \textbf{76.54} & 70.40 & 87.19 & 54.29 & 24.58 & 64.66 \\
\bottomrule
\end{tabular}}
\end{table*}

\subsection{Statistical Significance}
\label{subsec:stats-sig}

We report the precision of Table~\ref{tab:main} against the sampling noise of the evaluation: Table~\ref{tab:reliability} restates every cell of Table~\ref{tab:main} with a 95\% confidence interval. Each safety cell is a SAFE rate over $n$ held-out dialogues ($n=516$, $785$, $500$, $1400$, and $537$ for Adv., Harm., MB, CoS., and MHJ respectively; $3{,}738$ in total), with one exception: the \textit{Llama3.1-8B} Base MHJ cell is scored over $300$ dialogues, and its wider interval reflects that, so its sampling error is that of a binomial proportion; we report the 95\% Wilson interval as a half-width, and a behavior-clustered bootstrap interval for the pooled Avg. Over-refusal (XSTest, XSB) carries Wilson intervals ($n=250$, $340$), and MT-Bench a 95\% $t$-interval over its $160$ per-turn scores. Utility is the mean of six lm-eval tasks; since lm-eval reports an independent standard error per task, the mean carries a combined half-width of about one point.

Three trends follow from the intervals. First, the safety conclusions are resolved unevenly, and the intervals are worth reading before the point estimates. The clearest case is the self-verified reward on \textit{Qwen3-8B}: its average exceeds MUSE-D, the strongest baseline by average, by nearly nine half-widths, and its interval is disjoint from MUSE-D's on all five suites. That is a statement about MUSE-D specifically. On MHJ --- the hardest suite and the one most worth scrutinising --- the self-verified interval ($75.2\pm3.6$) overlaps MTSA's ($70.9\pm3.8$), so on that suite the two are not separated even though the point estimates differ by four points. The external reward on the same backbone is a different matter --- its average lead is under two half-widths and its per-suite intervals overlap the strongest baseline's on every suite, so on that backbone the two should be read as comparable rather than ordered. On \textit{Llama3.1-8B} both variants clear the strongest baseline on average, but the per-suite intervals overlap X-Boundary's on AdvBench for both, and on MultiBreak and MHJ for the self-verified variant. What the intervals support throughout is the separation between our variants and the \emph{field} of baselines, not a resolved ordering against the single strongest one on every suite. Second, the Utility column is indistinguishable across methods: every entry lies within roughly one combined interval of its base model, confirming that no method, ours included, meaningfully changes capability. Third, the helpfulness intervals counsel asymmetric reading: they are wide relative to the small method-to-method differences on XSTest and MT-Bench, which should therefore not be over-interpreted, whereas the elevated over-refusal of the aligned defenders relative to their base models clears the intervals comfortably and is a real cost rather than noise.

\begin{table*}[t]
\centering
\setlength{\tabcolsep}{4pt}
\caption{Table~\ref{tab:main} with 95\% confidence intervals, as \mbox{value\,{\scriptsize$\pm$half-width}}. Safety cells and over-refusal (XSTest, XSB) use Wilson intervals; the safety Avg.\ a behavior-clustered bootstrap; Utility a combined lm-eval standard error; MT-Bench a $t$-interval. Judge: \texttt{gpt-5-mini}.}
\label{tab:reliability}
\resizebox{\textwidth}{!}{
\begin{tabular}{@{}l cccccc c ccc@{}}
\toprule
& \multicolumn{6}{c}{\textbf{Safety} $\uparrow$} & \textbf{Utility $\uparrow$} & \multicolumn{3}{c}{\textbf{Helpfulness}} \\
\cmidrule(lr){2-7} \cmidrule(lr){8-8} \cmidrule(l){9-11}
\textbf{Method} & Adv. & Harm. & MB & CoS. & MHJ & Avg. & Avg. & XSTest$\,\downarrow$ & XSB$\,\downarrow$ & MT-Bench$\,\uparrow$ \\
\midrule
\multicolumn{11}{@{}c}{\textbf{\texttt{Llama3.1-8B-Instruct}}} \\
\midrule
Base          & 54.5\,{\scriptsize$\pm$4.3} & 62.5\,{\scriptsize$\pm$3.4} & 59.8\,{\scriptsize$\pm$4.3} & 84.9\,{\scriptsize$\pm$1.8} & 38.0\,{\scriptsize$\pm$5.4} & 59.9\,{\scriptsize$\pm$1.8} & 66.3\,{\scriptsize$\pm$1.0} & 10.0\,{\scriptsize$\pm$3.7} & 30.0\,{\scriptsize$\pm$4.8} & 6.20\,{\scriptsize$\pm$0.41} \\
SFT           & 69.4\,{\scriptsize$\pm$4.0} & 79.5\,{\scriptsize$\pm$2.9} & 87.0\,{\scriptsize$\pm$3.0} & 88.5\,{\scriptsize$\pm$1.7} & 55.1\,{\scriptsize$\pm$4.2} & 75.9\,{\scriptsize$\pm$1.4} & 65.0\,{\scriptsize$\pm$1.1} & 14.0\,{\scriptsize$\pm$4.3} & 45.6\,{\scriptsize$\pm$5.3} & 5.67\,{\scriptsize$\pm$0.39} \\
DPO           & 61.6\,{\scriptsize$\pm$4.2} & 71.5\,{\scriptsize$\pm$3.1} & 79.0\,{\scriptsize$\pm$3.5} & 88.3\,{\scriptsize$\pm$1.7} & 46.7\,{\scriptsize$\pm$4.2} & 69.4\,{\scriptsize$\pm$1.5} & 66.4\,{\scriptsize$\pm$1.0} & 10.4\,{\scriptsize$\pm$3.8} & 34.4\,{\scriptsize$\pm$5.0} & 6.26\,{\scriptsize$\pm$0.42} \\
SDGO          & 56.6\,{\scriptsize$\pm$4.2} & 61.7\,{\scriptsize$\pm$3.4} & 60.4\,{\scriptsize$\pm$4.3} & 85.0\,{\scriptsize$\pm$1.9} & 39.1\,{\scriptsize$\pm$4.1} & 60.6\,{\scriptsize$\pm$1.7} & 66.2\,{\scriptsize$\pm$1.0} & 11.6\,{\scriptsize$\pm$4.0} & 30.9\,{\scriptsize$\pm$4.9} & 6.11\,{\scriptsize$\pm$0.42} \\
MTSA          & 61.6\,{\scriptsize$\pm$4.2} & 71.6\,{\scriptsize$\pm$3.1} & 79.0\,{\scriptsize$\pm$3.5} & 86.6\,{\scriptsize$\pm$1.8} & 47.1\,{\scriptsize$\pm$4.2} & 69.2\,{\scriptsize$\pm$1.5} & 66.4\,{\scriptsize$\pm$1.0} & 10.8\,{\scriptsize$\pm$3.9} & 36.5\,{\scriptsize$\pm$5.1} & 6.12\,{\scriptsize$\pm$0.41} \\
MUSE-D        & 61.8\,{\scriptsize$\pm$4.2} & 67.5\,{\scriptsize$\pm$3.2} & 74.2\,{\scriptsize$\pm$3.8} & 87.4\,{\scriptsize$\pm$1.8} & 45.3\,{\scriptsize$\pm$4.2} & 67.2\,{\scriptsize$\pm$1.6} & 66.4\,{\scriptsize$\pm$1.0} & 8.4\,{\scriptsize$\pm$3.5} & 33.2\,{\scriptsize$\pm$5.0} & 6.27\,{\scriptsize$\pm$0.42} \\
X-Boundary    & 97.9\,{\scriptsize$\pm$1.3} & 95.0\,{\scriptsize$\pm$1.6} & 93.0\,{\scriptsize$\pm$3.0} & 93.6\,{\scriptsize$\pm$1.3} & 52.0\,{\scriptsize$\pm$4.2} & 86.3\,{\scriptsize$\pm$1.2} & 66.0\,{\scriptsize$\pm$1.0} & 16.8\,{\scriptsize$\pm$4.6} & 50.3\,{\scriptsize$\pm$5.3} & 6.26\,{\scriptsize$\pm$0.42} \\
\midrule
\multicolumn{11}{@{}l}{\textbf{Self reward}} \\
\quad -- unaligned & 63.0\,{\scriptsize$\pm$4.1} & 70.3\,{\scriptsize$\pm$3.2} & 64.2\,{\scriptsize$\pm$4.2} & 89.6\,{\scriptsize$\pm$1.6} & 44.3\,{\scriptsize$\pm$4.1} & 66.3\,{\scriptsize$\pm$1.7} & 66.3\,{\scriptsize$\pm$1.0} & 8.0\,{\scriptsize$\pm$3.4} & 33.8\,{\scriptsize$\pm$5.0} & 5.61\,{\scriptsize$\pm$0.40} \\
\quad -- aligned   & 98.8\,{\scriptsize$\pm$1.0} & 99.5\,{\scriptsize$\pm$0.5} & 96.6\,{\scriptsize$\pm$1.7} & 98.4\,{\scriptsize$\pm$0.7} & 58.5\,{\scriptsize$\pm$4.2} & 90.4\,{\scriptsize$\pm$1.0} & 66.8\,{\scriptsize$\pm$1.0} & 20.4\,{\scriptsize$\pm$5.0} & 54.4\,{\scriptsize$\pm$5.3} & 5.86\,{\scriptsize$\pm$0.39} \\
\multicolumn{11}{@{}l}{\textbf{External reward}} \\
\quad -- unaligned & 57.2\,{\scriptsize$\pm$4.2} & 65.2\,{\scriptsize$\pm$3.4} & 56.6\,{\scriptsize$\pm$4.3} & 88.5\,{\scriptsize$\pm$1.7} & 35.0\,{\scriptsize$\pm$4.0} & 60.5\,{\scriptsize$\pm$1.6} & 66.3\,{\scriptsize$\pm$1.0} & 8.8\,{\scriptsize$\pm$3.5} & 32.6\,{\scriptsize$\pm$5.0} & 5.81\,{\scriptsize$\pm$0.41} \\
\quad -- aligned   & 96.9\,{\scriptsize$\pm$1.5} & 98.2\,{\scriptsize$\pm$1.0} & 99.2\,{\scriptsize$\pm$0.9} & 99.0\,{\scriptsize$\pm$0.6} & 71.1\,{\scriptsize$\pm$3.8} & 92.9\,{\scriptsize$\pm$0.9} & 65.8\,{\scriptsize$\pm$1.0} & 21.6\,{\scriptsize$\pm$5.1} & 50.0\,{\scriptsize$\pm$5.3} & 5.21\,{\scriptsize$\pm$0.39} \\
\midrule
\multicolumn{11}{@{}c}{\textbf{\texttt{Qwen3-8B}}} \\
\midrule
Base          & 49.2\,{\scriptsize$\pm$4.3} & 52.5\,{\scriptsize$\pm$3.5} & 23.8\,{\scriptsize$\pm$3.7} & 82.6\,{\scriptsize$\pm$2.0} & 22.3\,{\scriptsize$\pm$3.6} & 46.1\,{\scriptsize$\pm$1.6} & 65.0\,{\scriptsize$\pm$1.0} & 7.2\,{\scriptsize$\pm$3.2} & 15.6\,{\scriptsize$\pm$3.9} & 6.08\,{\scriptsize$\pm$0.52} \\
SFT           & 59.9\,{\scriptsize$\pm$4.2} & 65.7\,{\scriptsize$\pm$3.4} & 49.6\,{\scriptsize$\pm$4.4} & 88.5\,{\scriptsize$\pm$1.7} & 35.6\,{\scriptsize$\pm$4.1} & 59.9\,{\scriptsize$\pm$1.7} & 65.9\,{\scriptsize$\pm$1.0} & 17.6\,{\scriptsize$\pm$4.7} & 40.9\,{\scriptsize$\pm$5.2} & 7.53\,{\scriptsize$\pm$0.33} \\
DPO           & 80.2\,{\scriptsize$\pm$3.4} & 86.6\,{\scriptsize$\pm$2.4} & 83.2\,{\scriptsize$\pm$3.3} & 97.5\,{\scriptsize$\pm$0.8} & 53.6\,{\scriptsize$\pm$4.2} & 80.2\,{\scriptsize$\pm$1.4} & 64.2\,{\scriptsize$\pm$1.0} & 26.0\,{\scriptsize$\pm$5.4} & 39.4\,{\scriptsize$\pm$5.2} & 5.87\,{\scriptsize$\pm$0.52} \\
SDGO          & 48.3\,{\scriptsize$\pm$4.3} & 52.5\,{\scriptsize$\pm$3.5} & 23.2\,{\scriptsize$\pm$3.7} & 82.4\,{\scriptsize$\pm$2.0} & 33.0\,{\scriptsize$\pm$3.9} & 47.9\,{\scriptsize$\pm$1.7} & 64.9\,{\scriptsize$\pm$1.0} & 10.0\,{\scriptsize$\pm$3.7} & 15.9\,{\scriptsize$\pm$3.9} & 6.23\,{\scriptsize$\pm$0.52} \\
MTSA          & 79.1\,{\scriptsize$\pm$3.5} & 90.2\,{\scriptsize$\pm$2.1} & 85.8\,{\scriptsize$\pm$3.1} & 97.8\,{\scriptsize$\pm$0.8} & 70.9\,{\scriptsize$\pm$3.8} & 84.8\,{\scriptsize$\pm$1.3} & 64.2\,{\scriptsize$\pm$1.0} & 38.4\,{\scriptsize$\pm$6.0} & 53.5\,{\scriptsize$\pm$5.3} & 5.99\,{\scriptsize$\pm$0.52} \\
MUSE-D        & 87.6\,{\scriptsize$\pm$2.8} & 90.8\,{\scriptsize$\pm$2.0} & 91.0\,{\scriptsize$\pm$2.5} & 97.4\,{\scriptsize$\pm$0.8} & 61.5\,{\scriptsize$\pm$4.1} & 85.7\,{\scriptsize$\pm$1.2} & 64.4\,{\scriptsize$\pm$1.0} & 20.4\,{\scriptsize$\pm$5.0} & 40.9\,{\scriptsize$\pm$5.2} & 5.87\,{\scriptsize$\pm$0.52} \\
X-Boundary    & 68.0\,{\scriptsize$\pm$4.0} & 71.1\,{\scriptsize$\pm$3.2} & 67.8\,{\scriptsize$\pm$4.1} & 88.8\,{\scriptsize$\pm$1.7} & 27.4\,{\scriptsize$\pm$3.8} & 64.6\,{\scriptsize$\pm$1.5} & 65.1\,{\scriptsize$\pm$1.0} & 18.8\,{\scriptsize$\pm$4.8} & 49.1\,{\scriptsize$\pm$5.3} & 5.69\,{\scriptsize$\pm$0.52} \\
\midrule
\multicolumn{11}{@{}l}{\textbf{Self reward}} \\
\quad -- unaligned & 69.0\,{\scriptsize$\pm$3.9} & 80.1\,{\scriptsize$\pm$2.8} & 66.8\,{\scriptsize$\pm$4.1} & 96.4\,{\scriptsize$\pm$1.0} & 46.9\,{\scriptsize$\pm$4.2} & 71.8\,{\scriptsize$\pm$1.6} & 65.0\,{\scriptsize$\pm$1.0} & 14.8\,{\scriptsize$\pm$4.4} & 37.9\,{\scriptsize$\pm$5.1} & 7.33\,{\scriptsize$\pm$0.36} \\
\quad -- aligned   & 97.1\,{\scriptsize$\pm$1.5} & 97.6\,{\scriptsize$\pm$1.1} & 97.6\,{\scriptsize$\pm$1.3} & 99.3\,{\scriptsize$\pm$0.4} & 75.2\,{\scriptsize$\pm$3.6} & 93.4\,{\scriptsize$\pm$0.9} & 64.6\,{\scriptsize$\pm$1.0} & 26.4\,{\scriptsize$\pm$5.4} & 56.8\,{\scriptsize$\pm$5.2} & 7.36\,{\scriptsize$\pm$0.34} \\
\multicolumn{11}{@{}l}{\textbf{External reward}} \\
\quad -- unaligned & 68.8\,{\scriptsize$\pm$4.0} & 77.3\,{\scriptsize$\pm$2.9} & 69.4\,{\scriptsize$\pm$4.0} & 95.5\,{\scriptsize$\pm$1.1} & 44.3\,{\scriptsize$\pm$4.2} & 71.1\,{\scriptsize$\pm$1.5} & 65.0\,{\scriptsize$\pm$1.0} & 10.0\,{\scriptsize$\pm$3.7} & 39.7\,{\scriptsize$\pm$5.2} & 7.20\,{\scriptsize$\pm$0.35} \\
\quad -- aligned   & 90.1\,{\scriptsize$\pm$2.6} & 93.6\,{\scriptsize$\pm$1.7} & 92.8\,{\scriptsize$\pm$2.3} & 98.4\,{\scriptsize$\pm$0.7} & 63.1\,{\scriptsize$\pm$4.0} & 87.6\,{\scriptsize$\pm$1.1} & 64.7\,{\scriptsize$\pm$1.0} & 16.8\,{\scriptsize$\pm$4.6} & 42.6\,{\scriptsize$\pm$5.2} & 6.96\,{\scriptsize$\pm$0.41} \\
\bottomrule
\end{tabular}}
\end{table*}
\subsection{Multiple Judges}
\label{subsec:more-judges}

\begin{table*}[h]
\centering
\scriptsize
\caption{Safety across five benchmarks under five judges held out from training. The corresponding \textit{gpt-5-mini} scores are in Table~\ref{tab:main}. \textbf{Bold}: best per column within each judge block.}
\vskip -0.1in
\label{tab:judge-safety-full}
\begin{tabular}{@{}l ccccc ccccc@{}}
\toprule
& \multicolumn{5}{c}{\textbf{\texttt{Llama3.1-8B-Instruct}}} & \multicolumn{5}{c}{\textbf{\texttt{Qwen3-8B}}} \\
\cmidrule(lr){2-6} \cmidrule(l){7-11}
\textbf{Method} & Adv. & Harm. & MB & CoS. & MHJ & Adv. & Harm. & MB & CoS. & MHJ \\
\midrule
\multicolumn{11}{@{}l}{\textit{Judge:} gpt-oss-20b} \\
\midrule
Base                   & 85.3 & 71.4 & 63.9 & 69.6 & 48.5 & 62.0 & 61.0 & 27.1 & 60.9 & 38.2 \\
SFT                                & 77.3 & 83.9 & 87.4 & 74.4 & 66.0 & 73.8 & 75.5 & 60.4 & 71.2 & 48.0 \\
DPO                                & 72.5 & 78.3 & 80.1 & 73.0 & 58.2 & 88.0 & 89.4 & 82.8 & 83.0 & 64.2 \\
SDGO                               & 66.0 & 70.2 & 62.7 & 68.5 & 52.1 & 62.2 & 61.5 & 28.2 & 61.9 & 44.7 \\
MTSA                               & 72.0 & 78.2 & 80.4 & 73.1 & 56.6 & 87.8 & 93.1 & 85.8 & 83.9 & 80.1 \\
MUSE-D                             & 70.1 & 75.4 & 75.0 & 71.7 & 56.7 & 92.6 & 94.4 & 90.6 & 85.4 & 71.5 \\
X-Boundary                         & 97.7 & 96.1 & 94.6 & 81.5 & 62.2 & 77.5 & 77.5 & 70.8 & 69.3 & 41.2 \\
\multicolumn{11}{@{}l}{\textbf{Self reward}} \\
\quad -- unaligned & 72.4 & 76.9 & 66.2 & 75.7 & 55.8 & 77.7 & 83.7 & 70.0 & 86.3 & 60.3 \\
\quad -- aligned   & \textbf{99.6} & \textbf{99.9} & 96.0 & 86.1 & 69.3 & \textbf{98.6} & \textbf{98.7} & \textbf{96.2} & \textbf{89.2} & \textbf{82.0} \\
\multicolumn{11}{@{}l}{\textbf{External reward}} \\
\quad -- unaligned & 69.2 & 73.0 & 59.3 & 71.0 & 49.6 & 77.9 & 82.9 & 71.8 & 87.1 & 55.6 \\
\quad -- aligned   & 97.9 & 99.2 & \textbf{98.2} & \textbf{90.9} & \textbf{79.9} & 94.8 & 96.3 & 92.4 & 86.0 & 72.6 \\
\midrule
\multicolumn{11}{@{}l}{\textit{Judge:} StrongREJECT} \\
\midrule
Base                   & 85.0 & 72.3 & 57.3 & 42.4 & 50.9 & 77.1 & 67.4 & 27.2 & 31.5 & 39.7 \\
SFT                                & 81.1 & 87.1 & 83.8 & 51.0 & 67.7 & 86.6 & 79.6 & 62.5 & 40.3 & 47.5 \\
DPO                                & 74.5 & 79.5 & 74.2 & 50.7 & 58.2 & 93.0 & 91.7 & 78.6 & 62.8 & 63.5 \\
SDGO                               & 70.4 & 72.9 & 57.9 & 44.3 & 53.2 & 77.5 & 66.6 & 27.0 & 31.4 & 45.7 \\
MTSA                               & 74.1 & 80.0 & 77.2 & 50.7 & 59.1 & 93.0 & 90.3 & 77.8 & 66.1 & 75.0 \\
MUSE-D                             & 74.4 & 76.9 & 72.1 & 48.2 & 58.1 & 94.8 & 93.6 & 85.0 & 66.2 & 71.1 \\
X-Boundary                         & 99.4 & 98.3 & 91.3 & 68.3 & 67.7 & 86.4 & 81.7 & 66.8 & 41.9 & 41.2 \\
\multicolumn{11}{@{}l}{\textbf{Self reward}} \\
\quad -- unaligned & 77.3 & 79.2 & 62.6 & 52.9 & 60.2 & 85.0 & 86.2 & 69.4 & 58.6 & 62.4 \\
\quad -- aligned   & 99.4 & \textbf{99.5} & 93.0 & 73.6 & 74.6 & \textbf{98.8} & \textbf{98.2} & \textbf{92.4} & \textbf{74.0} & \textbf{82.5} \\
\multicolumn{11}{@{}l}{\textbf{External reward}} \\
\quad -- unaligned & 74.6 & 77.7 & 58.4 & 45.0 & 56.4 & 90.9 & 86.8 & 73.9 & 55.7 & 62.3 \\
\quad -- aligned   & \textbf{99.6} & 99.4 & \textbf{96.8} & \textbf{77.4} & \textbf{84.9} & 98.6 & 96.6 & 87.6 & 67.0 & 74.6 \\
\midrule
\multicolumn{11}{@{}l}{\textit{Judge:} Llama-Guard-3} \\
\midrule
Base                  & 91.6 & 81.9 & 58.2 & 97.4 & 60.3 & 83.7 & 73.6 & 22.8 & 97.3 & 53.1 \\
SFT                                & 92.8 & 92.5 & 84.0 & 98.7 & 75.8 & 93.8 & 87.8 & 63.2 & 98.8 & 67.4 \\
DPO                                & 90.1 & 88.0 & 76.6 & 98.4 & 70.0 & 94.8 & 93.5 & 77.8 & 99.4 & 71.9 \\
SDGO                               & 84.7 & 80.3 & 58.8 & 97.3 & 66.3 & 85.1 & 75.8 & 20.8 & 97.4 & 52.9 \\
MTSA                               & 89.0 & 88.4 & 77.8 & 98.4 & 70.9 & 98.1 & 98.0 & 81.6 & 99.9 & 84.7 \\
MUSE-D                             & 88.4 & 84.3 & 72.0 & 97.9 & 68.9 & 98.3 & 96.6 & 86.2 & 99.4 & 76.0 \\
X-Boundary                         & 80.4 & 82.1 & 72.3 & 93.5 & 64.6 & 84.9 & 80.6 & 40.4 & 97.7 & 59.0 \\
\multicolumn{11}{@{}l}{\textbf{Self reward}} \\
\quad -- unaligned & 89.0 & 83.8 & 62.0 & 97.9 & 69.1 & 94.2 & 91.8 & 69.8 & \textbf{100.0} & 73.6 \\
\quad -- aligned   & \textbf{99.8} & \textbf{100.0} & 96.6 & \textbf{100.0} & 77.8 & \textbf{99.8} & \textbf{99.9} & \textbf{95.4} & \textbf{100.0} & \textbf{88.5} \\
\multicolumn{11}{@{}l}{\textbf{External reward}} \\
\quad -- unaligned & 87.6 & 82.5 & 57.8 & 97.9 & 62.9 & 95.2 & 92.7 & 74.8 & \textbf{100.0} & 74.5 \\
\quad -- aligned   & \textbf{99.8} & 99.5 & \textbf{99.8} & \textbf{100.0} & \textbf{85.3} & 99.4 & 99.5 & 93.8 & \textbf{100.0} & 82.5 \\
\midrule
\multicolumn{11}{@{}l}{\textit{Judge:} Qwen3Guard} \\
\midrule
Base                   & 65.3 & 46.2 & 61.0 & 90.1 & 47.3 & 42.4 & 40.8 & 26.0 & 93.1 & 44.3 \\
SFT                                & 52.9 & 61.9 & 88.0 & 91.7 & 66.3 & 52.5 & 55.5 & 61.4 & 96.5 & 53.1 \\
DPO                                & 50.8 & 58.5 & 79.2 & 93.2 & 58.7 & 69.0 & 75.9 & 84.4 & 98.1 & 60.9 \\
SDGO                               & 42.4 & 46.1 & 62.0 & 89.4 & 53.1 & 41.7 & 39.6 & 26.6 & 93.2 & 43.2 \\
MTSA                               & 50.0 & 58.6 & 80.0 & 92.9 & 58.5 & 74.8 & 83.8 & 87.4 & 99.3 & \textbf{81.8} \\
MUSE-D                             & 48.3 & 53.4 & 75.2 & 91.6 & 58.1 & 79.1 & 81.8 & 91.8 & 98.8 & 67.6 \\
X-Boundary                         & 36.8 & 53.2 & 73.3 & 82.4 & 52.3 & 43.8 & 47.0 & 42.8 & 93.5 & 48.4 \\
\multicolumn{11}{@{}l}{\textbf{Self reward}} \\
\quad -- unaligned & 49.0 & 53.4 & 64.4 & 90.0 & 53.3 & 47.9 & 58.1 & 67.6 & 97.1 & 56.1 \\
\quad -- aligned   & \textbf{98.4} & \textbf{98.6} & 98.6 & 99.1 & 69.6 & \textbf{90.1} & \textbf{91.1} & \textbf{97.8} & \textbf{99.9} & 78.8 \\
\multicolumn{11}{@{}l}{\textbf{External reward}} \\
\quad -- unaligned & 44.4 & 51.1 & 59.0 & 94.3 & 49.3 & 50.8 & 58.1 & 66.2 & 99.3 & 52.0 \\
\quad -- aligned   & 94.0 & 96.8 & \textbf{98.8} & \textbf{99.4} & \textbf{78.8} & 74.4 & 81.9 & 92.6 & 99.5 & 70.0 \\
\midrule
\multicolumn{11}{@{}l}{\textit{Judge:} HarmBench-13B} \\
\midrule
Base                   & 83.8 & 66.1 & 59.0 & 73.1 & 49.0 & 52.7 & 53.0 & 22.0 & 71.8 & 42.8 \\
SFT                                & 74.4 & 80.5 & 87.8 & 77.1 & 66.7 & 66.9 & 70.2 & 56.0 & 78.0 & 52.0 \\
DPO                                & 69.8 & 72.2 & 77.8 & 76.5 & 59.6 & 87.2 & 91.3 & 84.6 & 91.4 & 68.5 \\
SDGO                               & 65.1 & 65.0 & 61.0 & 73.5 & 52.5 & 49.4 & 54.9 & 20.8 & 71.6 & 49.7 \\
MTSA                               & 68.4 & 73.1 & 80.2 & 77.6 & 58.7 & 82.2 & 89.0 & 82.8 & 92.5 & 81.0 \\
MUSE-D                             & 69.6 & 70.6 & 75.4 & 75.9 & 57.4 & 94.2 & 95.4 & 90.4 & 92.5 & 76.4 \\
X-Boundary                         & 95.2 & 92.8 & 90.7 & 84.4 & 62.6 & 71.7 & 75.3 & 67.0 & 77.5 & 48.6 \\
\multicolumn{11}{@{}l}{\textbf{Self reward}} \\
\quad -- unaligned & 70.7 & 71.7 & 65.2 & 82.9 & 59.4 & 80.2 & 84.3 & 72.0 & 93.6 & 65.9 \\
\quad -- aligned   & \textbf{99.8} & \textbf{99.6} & 98.2 & 93.6 & 73.0 & \textbf{98.8} & \textbf{98.6} & \textbf{97.8} & 96.4 & \textbf{84.0} \\
\multicolumn{11}{@{}l}{\textbf{External reward}} \\
\quad -- unaligned & 67.4 & 69.0 & 57.0 & 84.3 & 52.7 & 77.3 & 84.1 & 75.6 & 95.0 & 63.5 \\
\quad -- aligned   & 98.6 & 98.7 & \textbf{99.6} & \textbf{99.0} & \textbf{82.5} & 97.9 & 97.3 & 96.2 & \textbf{96.9} & 74.7 \\
\bottomrule
\end{tabular}
\end{table*}

Table~\ref{tab:judge-safety-full} reports safety under five additional judges, all held out from training. The two rewards behave differently here, and the distinction matters. The self-verified reward places first or second in all 25 judge-by-benchmark cells on \textit{Qwen3-8B}, and is the strict maximum in 22 of them. The three exceptions are CoSafe under Llama-Guard-3, where all four of our rows tie at $100.0$; CoSafe under HarmBench-13B, where our external variant is marginally ahead ($96.9$ against $96.4$); and MHJ under Qwen3Guard, the one cell in the panel where a baseline leads it (MTSA, $81.8$ against $78.8$). The external reward is not comparably robust: it falls outside the top two in 9 of the 25 cells, ranking third or fourth on MHJ under every judge and third or fourth on AdvBench and HarmBench under Qwen3Guard. Claims about judge-independence should therefore be made for the self-verified reward specifically. The baselines also reorder among themselves --- MUSE-D leads MTSA under gpt-oss-20b, StrongREJECT and HarmBench-13B, and the two swap under Llama-Guard-3 and Qwen3Guard --- so what the judges agree on is the position of the self-verified reward relative to the field, not a total ordering. The lead is also narrower than the ranking implies: averaged over the five judges the self-verified reward exceeds the strongest baseline by $+7.1$ on MultiBreak but only $+2.6$ on MHJ and $+3.2$ on CoSafe, and on CoSafe under Llama-Guard-3 the margin is $+0.1$ with both our variants at $100.0$. The judges do differ in absolute calibration, with some scoring systematically stricter or more lenient than others, which is consistent with the moderate inter-judge agreement quantified in \S\ref{subsec:judge-corr}; crucially, however, these shifts move all methods together and leave the ranking essentially unchanged. Since our reward never consults any of these judges, this consistency indicates that the measured gains reflect genuinely safer behavior rather than overfitting to the idiosyncrasies of a particular evaluator.

\end{document}